\documentclass{article} 

\usepackage{arxiv}

\usepackage[utf8]{inputenc} 
\usepackage[T1]{fontenc}    
\usepackage{hyperref}       
\usepackage{url}            
\usepackage{booktabs}       
\usepackage{amsfonts}       
\usepackage{nicefrac}       
\usepackage{microtype}      
\usepackage{lipsum}

\usepackage{graphics} 
\usepackage{epsfig} 
\usepackage{mathptmx} 
\usepackage{times} 
\usepackage{amsmath} 
\usepackage{amssymb}  
\usepackage{multirow}
\usepackage{subfigure}

\usepackage{bm}
\usepackage{lscape}
\usepackage{array}
\usepackage{amsfonts}
\usepackage{amssymb}
\usepackage{color}
\usepackage{subfigure}
\usepackage[linesnumbered,lined,ruled]{algorithm2e}

\title{Understanding Human Context in 3D Scenes by Learning Spatial Affordances with Virtual Skeleton Models}

\author{
  Lasitha ~Piyathilaka  \\
  Center of Autonomous Systems\\
  University of Technology Sydney\\
  Australia \\
  \texttt{lasitha.piyathilaka@uts.edu.au} \\
   \And
  Sarath ~Kodagoda \\
  Center of Autonomous Systems\\
  University of Technology Sydney\\
  Australia \\
  \texttt{sarath.kodagoda@uts.edu.au} \\
}

\begin{document}

\maketitle
\thispagestyle{empty}
\pagestyle{empty}

\begin{abstract}
Robots are often required to operate in environments where humans are not present, but yet require the human context information for better human robot interaction. Even when humans are present in the environment, detecting their presence in  cluttered environments could be challenging. As a solution to this problem, this paper presents the concept of spatial affordance map which  learns  human context by looking at geometric features of the environment. Instead of observing  real humans to learn human context, it  uses virtual human models  and their relationships with the environment to map hidden human affordances in 3D scenes by placing virtual skeleton models in 3D scenes with their confidence values.  The spatial affordance map learning problem is formulated as a multi label classification problem that can be learned using  Support Vector Machine (SVM) based learners. Experiments carried out in a real 3D scene dataset recorded promising results and proved the applicability of affordance-map for mapping human context.

\end{abstract}


\section{INTRODUCTION}

Human Robot Interaction (HRI), which  is the study of interactions between humans and robots,  has gained significant attention over  recent years. This is due to the fact that  many robotics applications require  robots to work alongside  humans in a safe and acceptable manner, as opposed to  conventional robotics where robots operate in isolation. Therefore, it is believed that the ability of a robot to understand the human context in its surrounding environment is of paramount importance for a better human-robot interaction. 

In robotics, learning human context often involves tracking humans to learn their motion patterns \cite{sarathSocial}, human activity detection \cite{FSR,Gaussian}   and modeling relationships between humans and their surroundings \cite{Hallucinated3D}. Almost all of these techniques require robots to detect and track humans for a considerable amount of time before  being used to model human context. On the other hand, detecting, tracking and activity detection of humans in a cluttered environment  are still largely  open problems. Further, these tasks become more challenging and complicated when robots have to accomplish them while moving in a socially acceptable manner. Often these existing techniques require a  considerable amount of re-engineering when they are introduced into  new environments.

Introduced by Gibson in 1977 \cite{Gibson} affordance theory defines the word ``affordance"   as all  action possibilities latent in the environment, objectively measurable and independent of the individual's ability to recognize them \cite{Gibson,ChairChair,lasAffor}. Affordance theory argues that action possibilities are motivated by how environment is arranged. For example, chairs and sofas support the activity `sitting' and they are  physically designed  to support that affordance which  encourages the actor for sitting. Therefore, we believe this strong relationship between human context and environmental affordances could be used to learn human context even when humans are not observable. The rationale here is that, it is possible to learn environmental  affordances by   only looking at geometric features of the environment and this form the basis for the concept called affordane-map which is introduced in this paper. Affordance-map involves mapping possible human affordances in 3D scenes though virtual human models. This  affordance-map  learns the human context in a given environment without observing any real humans and bypasses challenges associated with human detection.

\begin{figure}
\includegraphics[width=4.5in]{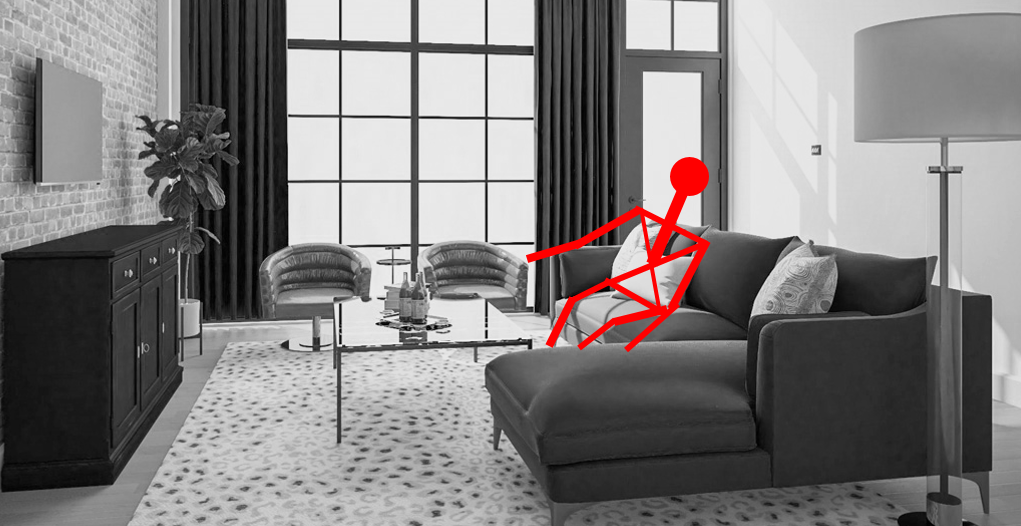}
\centering
\caption{ A 3D living room scene with virtual human models} 
\label{fig:human_context}
\end{figure}

The human context becomes `hidden' when humans are not observed in  a new environment. For example, consider the living room scene shown in Fig. \ref{fig:human_context}. The humans have an amazing ability to look at the scene  in  Fig. \ref{fig:human_context} and infer the human context in the environment.

If someone asked you to place a TV on this environment, you would probably place it in front of the sofas . For this you would imagine a few sitting humans and decide the final position of the TV based on the pose of these imagined humans so most of them could view the TV screen properly. If a service robot is required to locate the TV in this room, first it  could infer  sitting humans on sofas using their geometric features, and then it could use the human pose information to estimate the location of the TV. Such an approach would considerably reduce the robot's search space for the TV. This observation motivates us to learn the hidden human context in  3D scenes using the concept,`affordance-map', which involves mapping possible human affordances in 3D scenes though virtual human models. This  affordance-map  learns the human context in a given environment without observing any real humans and bypasses challenges associated with human detection.

As robots use grid based maps for localization, path planning and obstacle detection, affordance-map could   be used by a robot to improve the human robot interaction. Therefore, service robots operating in  indoor environments could largely benefit from learning the hidden human context. For example, a domestic service robot could use the human context information embedded in an affordance-map to arrange objects in a human preferred manner before humans arrive from work. Or, it could use pose information of virtual humans models embedded in affordance-map to search and localize various objects. Even when humans are present in the environment,  affordance-map could be used by a robot to carry out its task more efficiently. Firstly, it could use affordance-map to infer the possible human locations which could be used by the robot to efficiently detect humans. Secondly, it could use information from affordance-map to plan paths that minimize inferences to humans. Even affordance-map could provide strong priors for human activity detection when humans are partially observed or the views of them are completely obstructed.

\section{Related Work}

Automatic recognition of human context in an environment  is a well studied research area with a large body of previous work. Most of this previous work relies  on the fact that humans need to be detected and their activities need to be identified in order to successfully learn the human context in an environment. Therefore, most researchers have focused their studies on solving problems associated with human tracking and human activity detection.

One such popular approach is learning human context  by tracking human motion patterns using   various sensors installed in smart environments. These sensor networks include motion sensors, door sensors to detect human movements and sensors installed in equipment and cupboards \cite{Bang2008} \cite{Tapia2004}. The major challenge with such  systems is the requirement of large number of sensors and their adaptability to new environments. In addition, due to the low resolution and discrete nature of the data gathering the low level human context information  could be missed.

Various researchers have also focused on  the use of wearable sensors like gyros as a way of continuous human and activity detection \cite{Lara_asurvey}. However, the use of wearable sensors are too intrusive discouraging the applicability. Human context detection based on fixed video cameras have been heavily exploited in computer vision research. This includes human detection, tracking and human activity detection. However, in general, clutter contributes to low detection  accuracies \cite{Gupta2009}. In addition, video data can be obscured due to lighting conditions, type of costumes and background colors. Further, video  recordings raise privacy issues in many scenarios.

Some researchers have also  looked into the use of robotic systems for human activity recognition so the human context can be learned. The main advantage is  all sensors can be mounted on the robot so it can do active sensing.  In \cite{storkROMAN12}, audio-based human activity recognition using a non-markovian ensemble voting technique is presented. Although some of the human activities produce characteristic sounds from which the robot can infer activities, only some of the activities generate distinguishable sounds. Therefore, such a system may only  be used as a complement to the existing sensory systems. Recent advancements in video game consoles have invented low cost RGB-D cameras like Microsoft Kinect$^{TM}$. These cameras provide a wealth of depth information that a normal video camera fails to provide. In addition, RGB-D data from these sensors can be used to generate a skeleton model of a human with semantic matching of  body parts. These skeleton features have been used by several researchers to detect humans and their activities \cite{FSR, Gaussian}. The other main advantage of 3D sensing is location information of humans can be easily calculated. 

However, learning human context by watching real humans could become problematic in most scenarios. First, robots will have to observe humans for a considerable  period of time before learning the human context in a new environment. On the other hand, many robots are often required to operate in environments where real humans would not be observed at all, but robots still need the human context information to carry out various tasks. A good example of this is a situation where a service robot is required to arrange small items in a house before humans arrive home from work. In this scenario, the robot needs human context information to arrange objects in a human preferred manner and would have to gather human context information without observing real humans. 

Even though humans are visible, learning human context from them could become problematic in most scenarios. For example, learning human context could become challenging when the human  subject moves away from the sensory range or when the sensor inputs become obscured. One way to fix this problem is to mount sensors on a mobile robot and guide the robot towards humans. However, when the robot starts to move it could create several other social issues. For example,  the robot might  interfere with the activity that the human is performing or it could collide with the human if the possible motion path of the human is blocked. The other major issue with observing real humans is privacy of humans.  It is no doubt that the presence of a robot inherently affects a human's sense of privacy \cite{Kahn06whatis}. Privacy relates to the right of an individual to decide what information about himself or herself can be shared with others. Research in the field of human-computer interaction has shown that the user should have sufficient level of awareness and control of shared information in order to achieve widespread acceptance of new technologies \cite{Bellotti93designfor}. Therefore, when sensors like cameras are used to learn human context privacy issues could become paramount.

Because of problems associated with human detection and tracking researchers have tried various other techniques that could be used for learning human context. One such popular method, the concept of affordance  has become a focus of attention within the cognitive vision and robotics community lately.
Psychologist James J. Gibson originally introduced affordance in his 1977 article `The Theory of Affordances' \cite{Gibson} and explored it more fully in his book `The Ecological Approach to Visual Perception' \cite{gibson2014ecological} in 1979. The  Affordance theory states that the world is perceived not only in terms of object shapes and spatial relationships but also in terms of object possibilities for action (afordances). In other terms,  perception drives actions. Therefore,  it could be argued that human action possibilities and thus the human context could be inferred by looking at the environmental properties. Further in his book ``The Ecological Approach to Visual Perception'' he  explains `sittable' affordance as follows.

\begin{quote}
``
Different layouts afford different behaviors for different animals, and different mechanical encounters. The human species in some cultures has the habit of sitting as distinguished from kneeling or squatting. If a surface of support with the four properties (horizontal, flat, extended and rigid) is also knee high above the ground, it affords sitting on. We call it a seat in general, or a stool, bench, chair, and so on, in particular. It may be natural like a ledge or artificial like a couch. It may have various shapes, as long as its functional layout is that of a seat. The color and texture of the surface are irrelevant. Knee high for a child is not the same as knee-high for an adult, so the affordance is relative to the size of the individual. But if a surface is horizontal, flat, extended, rigid, and knee high relative to a perceiver, it can in fact be sat upon. If it can be discriminated as having just these properties, it should look sit-on-able. If it does, the affordance is perceived visually. If the surface properties are seen relative to the body surfaces, the self, they constitute a seat and have meaning'' - \bf{Gibson 1977}, \textit{The Ecological Approach to Visual Perception}.
\end{quote}

Motivated by these arguments in affordance theory,  a few researchers have recently attempted to learn action possibilities hidden in the environment by only looking at geometric properties. These approaches bypass the need for human detection and tracking.

In \cite{ChairChair}, researchers use virtual human models to recognize objects that have  `sittable' affordance without using common approaches such as 3D features for object recognition. They use a human model with a sitting pose to calculate distance features to learn an affordance model, which is modelled as  a probability density. These affordance probabilities are calculated for each 3D grid location of the room and the locations with highest probabilities are classified as 'sittable'. The affordances model is trained on a  synthetic dataset with different `sittable' furniture types. Although they  achieved good recognition accuracies with synthetic datasets, they  failed to record good results when tested on a real 3D environment.

Although the  results of \cite{ChairChair} are encouraging, it has the following limitations when applied for affordance mapping. First, the learning process is not discriminative. Therefore, it cannot distinguish the difference between `sittable' and 'non-sittable' locations correctly. Because of this, it would classify the locations with high prediction probability as `sittable'  in a room where any `sittable' locations are not observed. On the other hand, in this method parameters of the training model are trained using furniture downloaded from a synthetic dataset and their arrangements  in real rooms are not considered. However in real world scenarios, the distance features easily  could be affected by nearby objects, walls and the floor. Perhaps this could be the major reason for this algorithm  recording  high false positive detections in  real 3D scenes. 

Recently, a few researchers introduced virtual humans models to learn human object relationships (Affordances)  in 3D scenes \cite{jiang2012learning}. First, object human relationships are learned from labeled objects. Then learned models are used to arrange objects in a human preferred manner in a synthetic environment. However, they assumed that the objects have been detected first before modelling affordances and therefore cannot be used in a new 3D room without object labels. As an extension for this research, in \cite{jiang2013hallucinated} researchers propose to use Infinite Factored Topic Model (IFTM) to model object human relationships and use them to improve the object detection accuracy. They used hallucinated humans to derive object human relationship features. However, object locations and locations of human models are not initially known but learned jointly from environmental features during the training process.  However, the main intention of this method is to improve 3D object detection accuracy rather than correctly learning the correct location of the human model. Therefore, locations of the human models inferred by this method might not be optimum  and heavily dependent on the presence of object types that were used for training affordance models.

The affordance mapping process presented in this paper differs from these existing algorithms in three ways. Firstly, the proposed affordance learning process is formulated as a multi-label binary classification problem. Therefore, unlike existing methods, it can correctly label each grid locations of the 3D scene either with a positive label or negative label with a confidence value. Secondly, object human relationships are modelled implicitly and are included as features in the training phase. Therefore the model is not required to detect objects or their labels to learn the affordance-map. Thirdly, this thesis proposes a framework to build an affordance-map in a large room efficiently with multiple types of affordances

\section{Mapping Affordances}
\begin{figure}[ht]
\centering
\subfigure[Vitual Skeleton Map]{%
\includegraphics[width=2.5in]{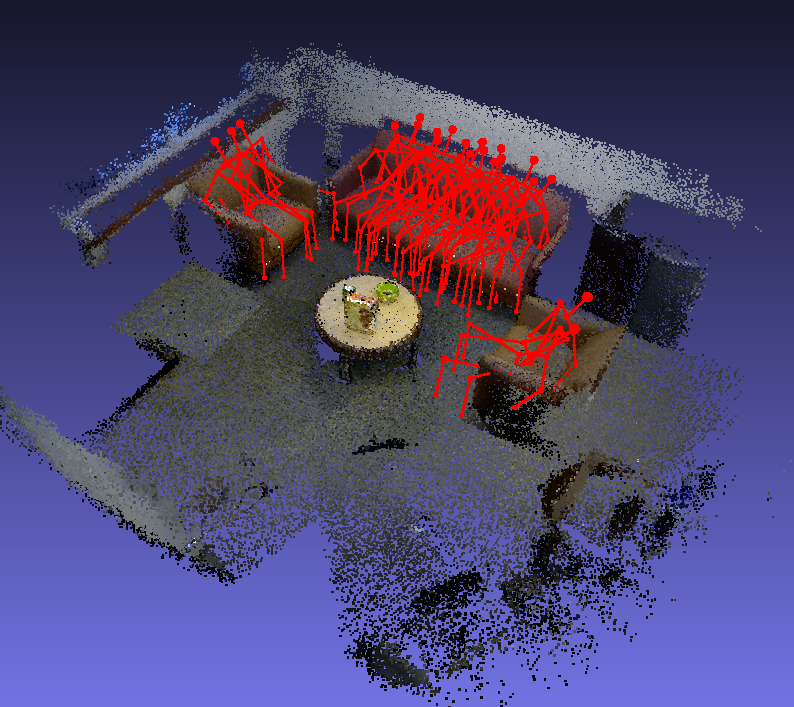}
\label{fig:skeletons_intro}}
\quad
\subfigure[The contour map of affordance likelihoods]{%
\includegraphics[width=2.5in]{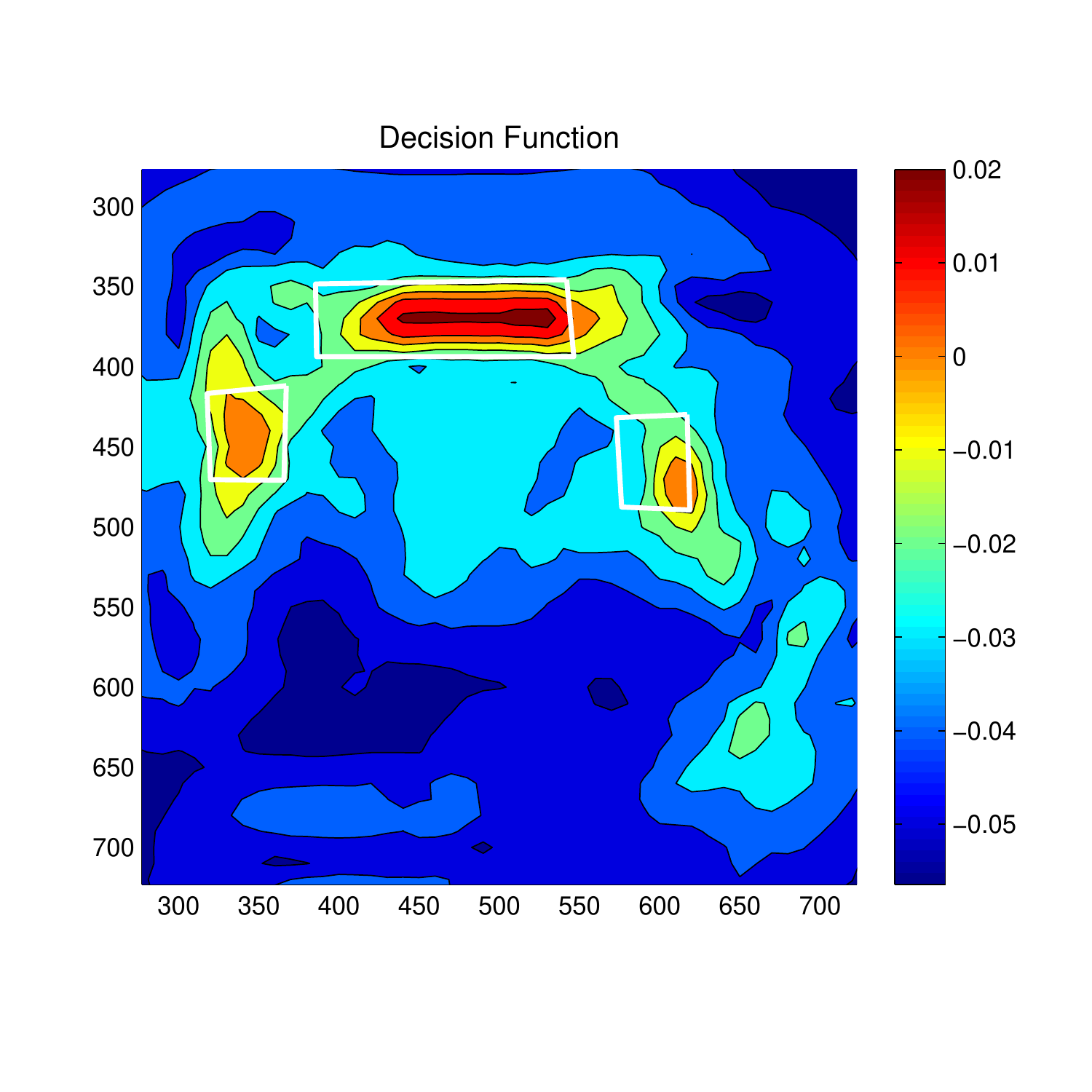}
\label{fig:decison_intro}}
\caption{The affordance-map of a living room}
\label{fig:affor_map_intro}
\end{figure}

Affordance-map is consisted of two sub-maps as shown in  Fig. \ref{fig:affor_map_intro}. First, it predicts virtual human skeleton models in locations that support the tested affordance as shown in  Fig. \ref{fig:skeletons_intro}. Then it outputs a confidence value for each  predicted skeleton that indicates how likely the skeleton model is being supported by the surrounding environment as shown with a contour map in Fig. \ref{fig:decison_intro}.

The affordance-map building process can be formulated as a supervised learning problem. Given a set of 3D pointcloud images $ \lbrace i_1,....,i_k\rbrace \subset I$ and their associated affordance-map  \textit{A}, the goal is to learn mapping $ g:I \rightarrow A$ which can be later used to automatically build affordance-map in unseen pointclouds. As some  locations of the map could have multiple affordances-labels, mapping $ g:I \rightarrow A$  becomes a multi label classification problem. As affordance types are not mutually exclusive, one way to simplify this problem is to divide each 3D image space into four dimensional grid locations ( $(x,y,z)$ location of the skeleton model and $\theta$ orientation) and  build an independent  binary classifier for each affordance type. Each binary classifier predicts a binary label vector $\bar{\mathbf{y}}_k=(y_1,.y_i..,y_n)$ where   $y_i \in\lbrace +1, -1\rbrace $  for the feature vector   $\mathbf{\bar{x}}_k= (x_1,....,x_n)$,  $x_i \in \Re ^n$ calculated for each  grid location  of the 3D scene. The label $y_i$ becomes $+1$ if that location support the tested affordance and $-1$ if not.

\section{Supervised Learning for Affordance Detection}

Binary classification forms the basis for  the concept of affordance-map. Since affordance labels are not mutually exclusive, affordance mapping can be  categorized  as a multi-label classification problem. Given $n$ number of feature vectors $\bar{\mathbf{x}}_k= (\mathbf{x}_1,....,\mathbf{x}_n)$  for each grid location of the $k^{th}$  3D image, this  classifier tries to assign $n$ number of affordance label vectors, $\bar{\mathbf{a}}_k=(\mathbf{a}_1,....,\mathbf{a}_n)$  to $n$ grid locations of the image. 

\begin{figure}
			\includegraphics[width=4.5in]{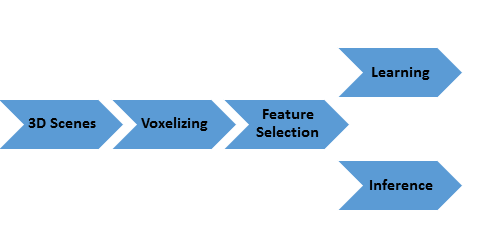}
			\centering
			\caption{Steps in learning and inference stages}
			 \label{process}
			 \end{figure}

Fig. \ref{process} summarizes the learning and inference process. Both learning and inference stages first require to voxelise the 3D scene into grid locations as explained in  section \ref{sec_vox}. Then 3D features are calculated for each and every grid location as explained in section \ref{sec_features}. These features and class labels from the dataset are later used to learn parameters of the classification model. 

In inference stage, skeleton models are moved across every grid location with all possible orientations while calculating features for each and every  grid location. These features and learned parameters of the classification model are then used to predict labels for the new 3D scene.

\subsection{Voxelising 3D Scenes}
\label{sec_vox}

The affordance map is a concatenation of classifier's outputs for each $(x,y,z,\theta)$ grid location of the  3D scene. Therefore, the first step of the affordance mapping involves voxelising the input 3D images into grid locations. This voxelising is done by dividing the input image into  10 cm x 10 cm x 10 cm grids. The rotation $\theta$ of each skeleton model is evaluated at 0.1 rad resolution at each grid level. In order to limit the search space, the grid search  along the $z$ axis can be restricted  as the human skeletons are always located close to the ground plane. Therefore, the height to the Torso from the ground plane can be estimated and the grid search along the z axis can be  restricted  to one grid level above and below the torso position. Even with these simplifications, the search space for a 10m x 10m room could extend up to  1,890,000 (100 x 100 x 3 x6 3) unique grid locations

\subsection{Feature selection}
\label{sec_features}

The proposed  affordance-map predicts virtual skeleton models with likelihood values for each $(x_i,y_i,\theta_i)$ grid location of the  3D image. It is based on a  binary classifier that  predicts  positive labels to  the locations that  support the tested affordance and negative  labels  to the locations that do not support the tested affordance. Therefore, the classifier's performance largely depends on the types of features used. These features should be highly informative such that the classifier would be able to  predict class labels correctly.

In this thesis, a new set of features  based on virtual human skeleton models is proposed  for mapping affordances. These features directly model the relationship between the humans and the environment. Following sections describe different types of skeleton models and their associated features used by binary classifiers to build the affordance-map.

\subsubsection{Virtual Human Skeleton Model} 

Instead of observing real humans in the environment, the proposed affordance mapping process uses virtual humans  to model interaction between the environment and the humans.  Although  many  human poses could be observed in a given  environment, very few of them directly  influence the context of the environment. For example, the most frequently observed human pose in an office environment is sitting and working at an office desk. Therefore,  if the locations within the office room that support this affordance can be identified then the human context of the environment can be inferred easily.

\begin{figure}[ht]
\centering
\subfigure[Sitting-Relaxing]{
\includegraphics[width=1.9in]{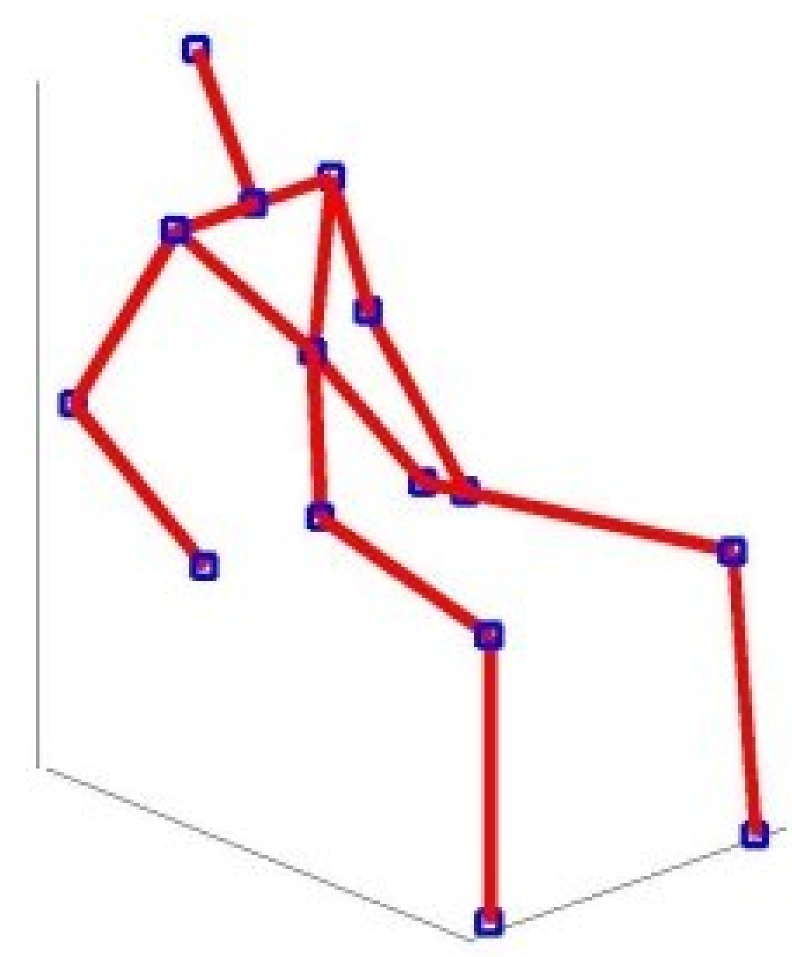}}
\quad
\subfigure[Standing-Working]{
\includegraphics[width=1.5in]{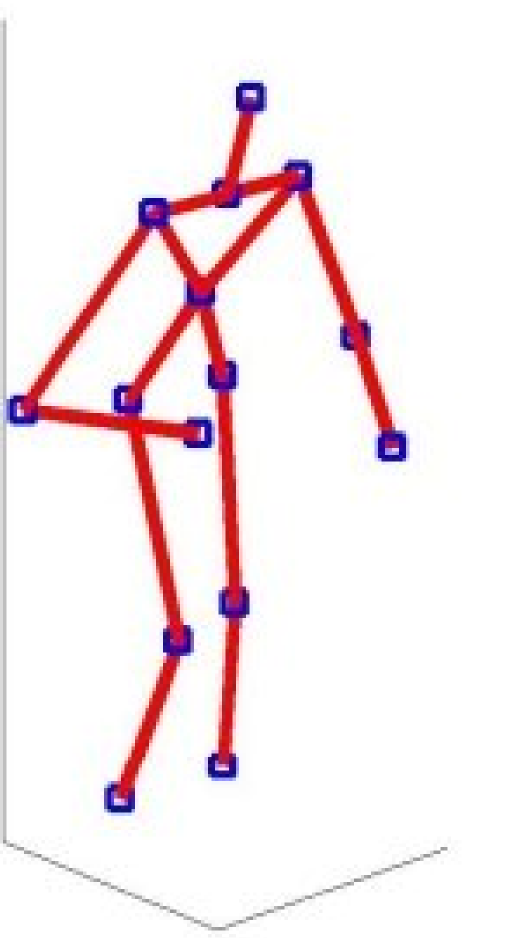}}
\subfigure[Sitting-Working]{
\includegraphics[width=1.5in]{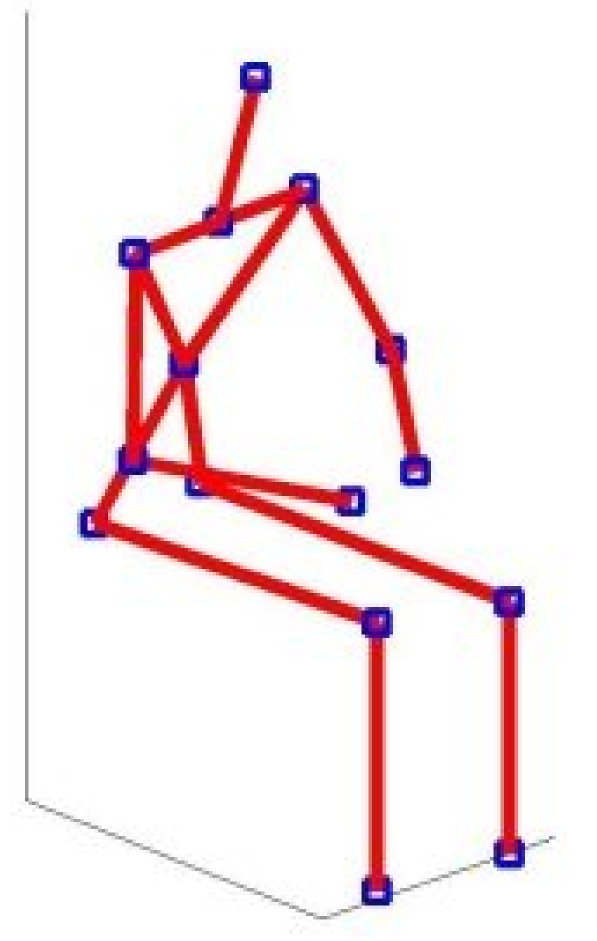}}
\caption{Types of affordances and their associated skeleton models }
\label{fig:pose_affordances}
\end{figure}

For the purpose of mapping affordances, human skeleton models are obtained from a  human activity detection dataset \cite{c1}. These human skeleton models are captured using a depth camera from real humans while they perform different activities. The K-mean clustering classifies all skeleton models into three clusters and the most frequently seen pose in the  each cluster is shown in the Fig .\ref{fig:pose_affordances}. Each of these skeleton models are associated with an affordance type that closely represents the type of activity that each skeleton model exhibits.

Each human model  is a  skeleton with 15 joint body positions in 3D. In order to obtain feature vectors, these virtual skeleton models need to be transformed to each $(x,y,z,\theta)$ pose  of the map. This can be done by moving the 3D points of the  human skeleton, $H_{l}$ across given environment using the rigid body transformations of translation and rotation. Then  each human skeleton model can be mapped to the coordinate system of the environment using (\ref{eqskel}), where $\mathbf{g}_k=(x_k,y_k,z_k,\theta_k)$ is the position and orientation of  the skeleton's torso in the world coordinate system and $R_{z}(\theta_k)$ is  the rotational matrix about the $z $ axis (vertically up). It is to be noted that only rotation about the $z$ axis is considered here.

\begin{equation}
H_{w}(\mathbf{g}_k)=[x_k,y_k,z_k]^T+R_{z}(\theta_k).H_{l}
\label{eqskel}
\end{equation}

\subsubsection{Distance and Collision Features}

The features  model the relationship between the human skeletons and the environment. This relationship is modeled  through two geometric features; distance features and collision features. Selection of these features are motivated by two facts. First is the proximity of objects for effective interactions and the second is to prevent collisions with objects of the environment. 

\begin{figure}
			\includegraphics[width=2.5in]{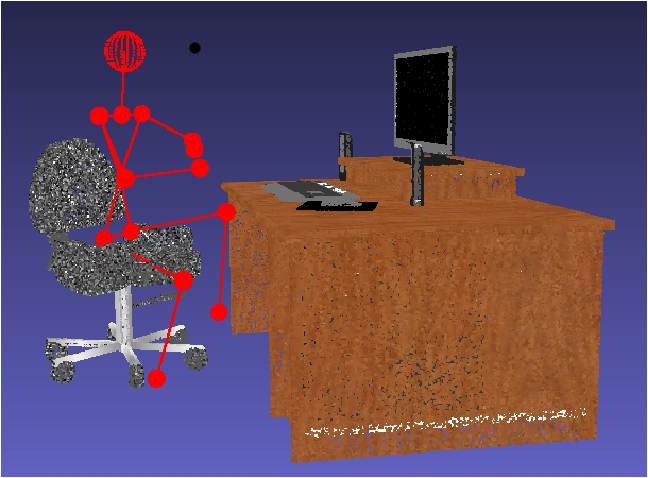}
			\centering
			\caption{Virtual human models sitting near a computer}
			 \label{poses2}
			 \end{figure}

The virtual humans and their interactions with the environment  can be illustrated by the example shown in  Fig. \ref{poses2}. It can be seen from   Fig. \ref{poses2} that proper sitting of the virtual human model can be described by placing the spine supporting the vertical part of the chair while thighs supporting the horizontal part of the chair. Also the human model should  not go through any 3D point of the chair. Therefore, if each point of the chair is converted in to a 3D distance field then each point of the virtual human model should lie with a specific distance value from the chair.

The first step of feature extraction involves modeling the environment. The 3D point clouds generated from  RGBD SLAM algorithms \cite{ccny}  usually contain a large number of 3D points, and searching for a particular affordance in this large feature  space is computationally infeasible. Therefore, it is required to convert these dense point clouds into much lighter abstract representations without losing much data.  This is achieved by  modeling the environment with a  3D Distance Transform Map $DT(\textbf{x})$ and a 3D Occupancy Map $OC(\textbf{x})$, where $\textbf{x}$ is any 3D position of  the environment. The 3D Distance Transform (DT) is a  shape representation that indicates the minimum distance from a point in the environment to the closest occupied voxel. In this approach, the 3D Distance Transform is calculated by using the occupied voxels  of 3D point clouds,  $OC$. 
 The distance transform map $DT(\textbf{x})$ of the occupancy grid map $OC$ can be  generated using an unsigned distance function (\ref{eqDis}), that represents the Euclidean distance from each  location $\textbf{x}$ of the environment  to the nearest  occupied voxel in $OC(\textbf{x})$.
\begin{equation}
DT(\textbf{x})=\underset{O_{j} \in  OC}{min}|O_{j}-\textbf{x}|
\label{eqDis}
\end{equation}

The distance transform map given by the the equation (\ref{eqDis}) can be calculated very efficiently, but yet can provide an informative representation of the dense input  3D pointcloud.

The distance features are obtained by moving the human model across the voxels in the environment and calculating a distance measure for each and every skeleton points of the human model. Once the environment is  modelled by (\ref{eqDis}), we can effectively calculate distance features of a human skeleton with location and orientation  $\mathbf{g}_k =(x_k,y_k,z_k,\theta_k)$ by (\ref{features}), where $n$ is the number of 3D points in the skeleton.

\begin{equation}
[d_1, d_2,... d_n] =DT(H_{w}(\mathbf{g}_k))
\label{features}
\end{equation}

Similarly, we can check  for any collisions of a skeleton at any location and orientation, $X_k $ by (\ref{collison}). In case of a collision  $c_i$ is assigned as 1 and 0 otherwise.

\begin{equation}
[c_1, c_2,... c_n] = OC(H_{w}(\mathbf{g}_k))
\label{collison}
\end{equation}

\subsubsection{Normal Features}

The other set of features used for affordance detection is normal features. These normal features represent vertical and horizontal planes of the environments. The selection of these features is motivated by the fact that most of the affordances are supported by vertical and horizontal planes. For example, `sitting' affordance is supported by a horizontal plane under the lap of a sitting skeleton and the spine of the skeleton is supported by a vertical plane. 

\begin{figure}
\includegraphics[width=2.5in]{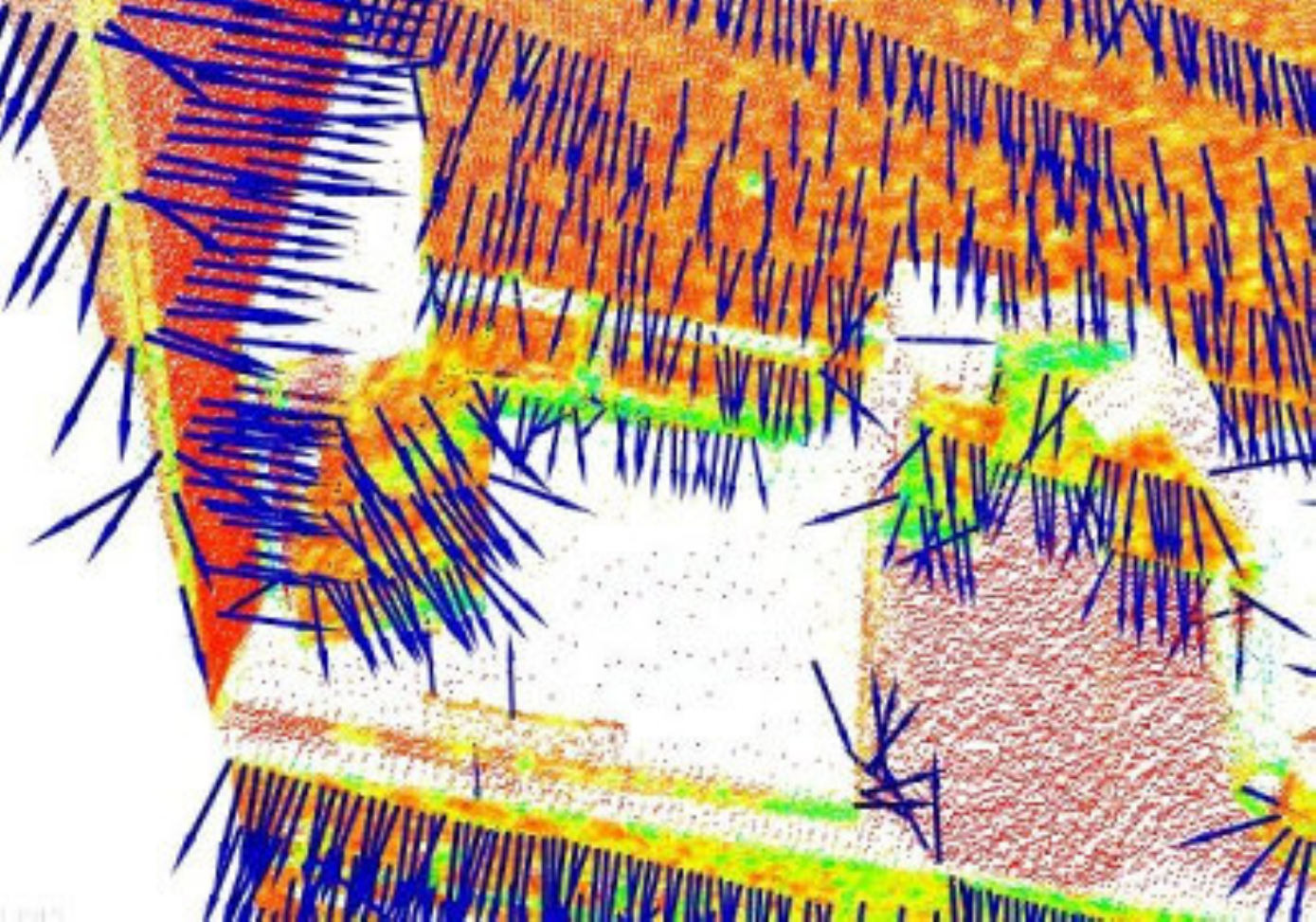}
\centering
\caption{Surface Normals calculated from point clouds}
\label{fig:normals}
\end{figure}

Given a geometric surface, it is usually trivial to infer the direction of the normal at a certain point on the surface as the vector perpendicular to the surface in that point. However, since the point cloud data only contains data points, calculation of surface normals involves two steps. First, surface meshing techniques are used to  obtain the underlying surface from the acquired point cloud data. Then surface normals are calculated from this mesh.  Fig. \ref{fig:normals} shows surface normals calculated from a set of point-cloud data.

The normals features for the affordance detection are calculated as follows. First, a 1m x1m x 1m cubic volume is considered from the torso position of  a skeleton model at location, $(x,y,z,\theta)$. Then it is voxelised into 10cm x 10cm x 10cm voxels. Finally surface normal values of points in each grid cell are averaged to find the  normal features for each voxel. Normal features alone could be highly sensitive to observation noise and could affect the classification accuracy. However, averaging normal features over each grid location filters out sensor noise and improves the classifier's performance.

\section{Experimental Setup}

This section explains the experimental setup used for mapping affordances in 3D pointclouds.  

\subsection{Dataset}
Both the  SVM and  Flexible Naive Bayes  classifier discussed in previous sections are supervised learning algorithms. Therefore, they need to be trained first before used for  building an affordance-map in an unseen environment and require a number of 3D scenes to learn parameters of the classification models. The recent advancements in RGBD Simultaneous Localization and Map Building (SLAM) algorithms (\cite{ccny}) allows us to build highly informative  dense 3D scenes by stitching  3D point clouds frames acquired from low cost 3D depth cameras. A dataset is created for affordance mapping by  capturing  a number of dense 3D scenes using a ASUS Xition depth camera and a 3D map building software \cite{ccny}. The final dataset is consisted of  12 high quality 3D scenes captured in  office and domestic environments.

\subsection{Search space}

The affordance map is a concatenation of classifier's outputs for each $(x,y,z,\theta)$ grid location of the  3D scene. Therefore, the first step of the affordance mapping involves voxelising the input 3D images into grid locations. This voxelising is done by dividing the the input image into  10 cm x 10 cm x 10 cm grids. The rotation $\theta$ of the each skeleton model is evaluated in 0.1 rad resolution at each grid level. In order to limit the search space, the grid search  along the $z$ axis can be restricted  as the human skeletons are always located close to the ground plane. 

\subsection{Ground Truth Labels}
\label{sec_Ground_Labels}
The proposed affordance-map building process is a  supervised learning problem, which requires ground truth labels in order to learn parameters of the classifier. Therefore, all  possible locations in 3D images  that support the tested affordances are manually labeled for each affordance type.

\section{SVM Classifier for Affordance-Detection}

Support Vector Machines (SVMs) \cite{cortes1995support,joachims1999making,suykens1999least} is a popular machine learning technique, which has been successfully applied to many real-world application domains. The goal of the SVM learning algorithm is to find the optimal separating hyperplane which effectively separates instances into two classes. SVM algorithm is a discriminative binary classifier and it's decision function can be used to assign confidence values for its predictions. These reasons make the  SVM classifier  an ideal candidate for mapping affordances. However the nature of the affordance detection problem posses some unique challenges as described in following sections.

\section{Soft Margin Optimization with Class Imbalance}
The Table. \ref{tb:imbalance_data} summarises the number of positive and negative examples found in the dataset for each affordance type. It is clear form this table data that all of the affordance types found in the dataset  have a high  number of negative examples than the number of  positive examples. The sitting-working affordance has recorded the highest class imbalance. This is desirable as the sitting surface of an office chair is very small with compared to the rest of the room.  Any  affordance detection algorithm  should be able to effectively  handle these type of class imbalances in order to effectively build affordance map in a large room. However, a few researchers have reported that SVM classifier tend to produce sub-optimum results when trained on highly imbalanced datasets \cite{tang2009svms,akbani2004applying}. 

\begin{table}[]
\centering
\caption{Summary of the Class Imbalance in the Dataset}
\label{tb:imbalance_data}
\begin{tabular}{|l|c|l|l|l|c|l|}
\hline
\multicolumn{1}{|c|}{\multirow{2}{*}{{\bf Affordance}}} & \multicolumn{4}{c|}{\multirow{2}{*}{{\bf \begin{tabular}[c]{@{}c@{}}\# Positive \\ Examples\end{tabular}}}} & \multirow{2}{*}{{\bf \begin{tabular}[c]{@{}c@{}}\# Negative \\ Examples\end{tabular}}} & \multirow{2}{*}{{\bf \begin{tabular}[c]{@{}l@{}}Imabalance \\ Ratio\end{tabular}}} \\
\multicolumn{1}{|c|}{}                                  & \multicolumn{4}{c|}{}                                                                                       &                                                                                       &                                                                                    \\ \hline
Sitting-Relaxing                                        & \multicolumn{4}{c|}{2457}                                                                                   & 1285326                                                                               & 1:523                                                                              \\ \hline
Sitting-Working                                         & \multicolumn{4}{c|}{213}                                                                                    & 2570439                                                                               & 1:12067                                                                            \\ \hline
Standing-Working                                        & \multicolumn{4}{c|}{391}                                                                                    & 1284935                                                                               & 1:3286                                                                             \\ \hline
\end{tabular}
\end{table}

\begin{figure}
 \centering
\includegraphics[width=4.5in]{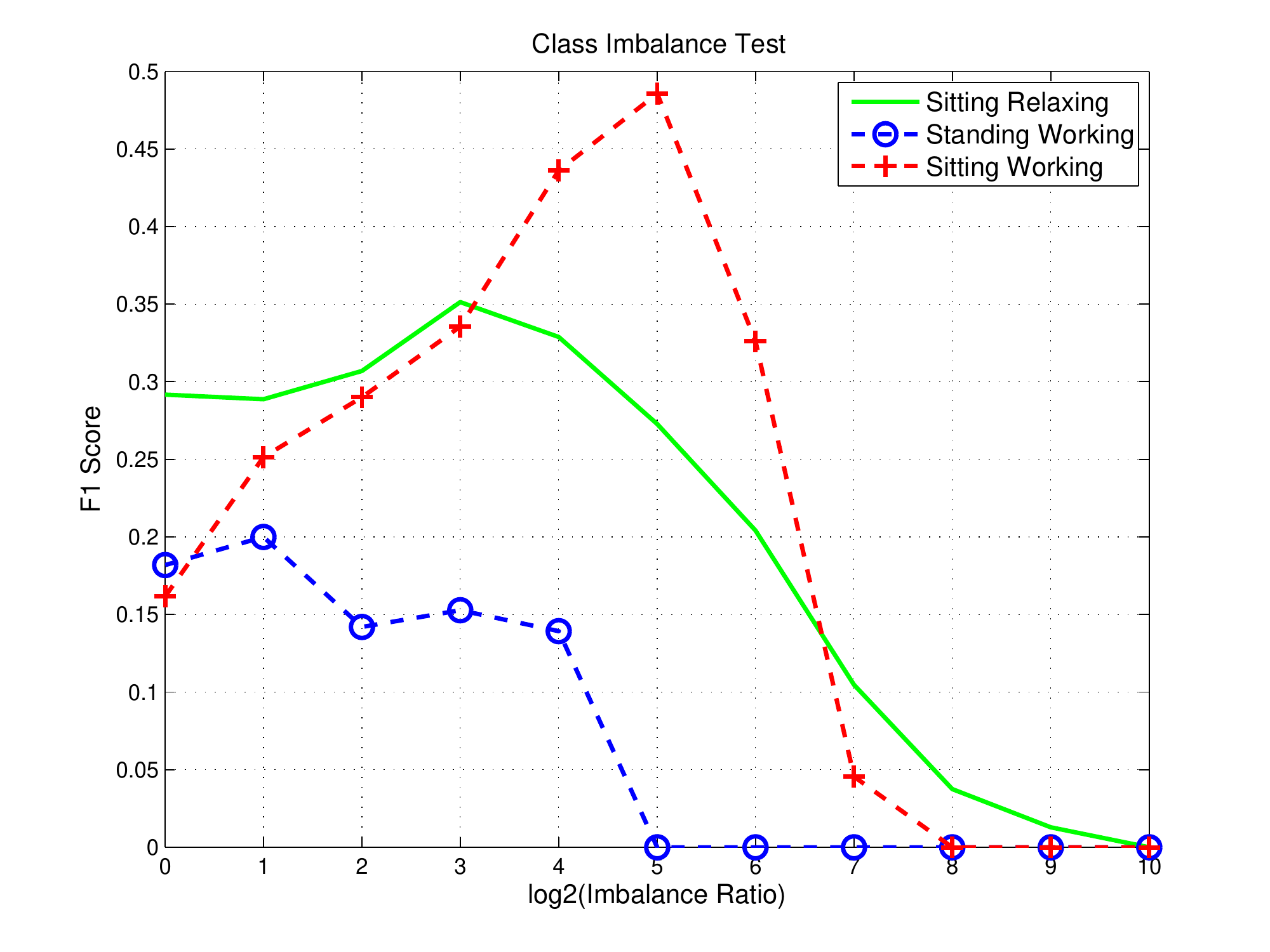}  
  \caption{The effect of class imbalance on the SVM classifier} 
  \label{fig:random_imbalance}
\end{figure}

To understand the impact of the class imbalance on affordance detection, we carried out few experiments and experimental results are shown below. First the dataset is divided into two subsets: training set and validation set. Then SVM classifiers are trained for different class imbalances. These learnt classifiers are  tested on  unseen 3D images and F1-scores are recorded for each setting. The average F1-scores for each affordance  type in different class imbalances are shown in  Fig. \ref{fig:random_imbalance}.

According to the test results, it is clear that  all three affordances have reported low F1- scores initially (when the positive and negative classes are equally balanced) and increased slightly  before gradually dropping down again. This behaviour is clearly visible with the sitting-working affordace and less visible with the standing-working affordance.  When more negative examples are added for training, the performances of the SVM classifiers have gradually increased as can be seen with increasing F1-scores.  This is predicted because more negative examples mean the SVM classifier can discriminate the difference between the positive examples and negative examples more effectively. However, the performances of the SVM classifiers have degraded when the class imbalances become more and more extreme. When the class imbalances are too high all  F1-scores have fallen to zero. That means the  classifier have predicted all locations of the room with negative labels. Overall, standing-working affordance has recorded low F1-scores. This unexpected behaviour is due to the movement of the separating hyperplane of the SVM classifier toward the majority class and it can be  be understood by analysing the  SVM soft-margin optimization problem. 

\begin{equation}
\begin{split}
\arg\min_{(\mathbf{w},b)}\lbrace\frac{1}{2}\|\mathbf{w}\|^2 + C \sum_{i=1}^n \xi_i \rbrace\\
y_i(\mathbf{w}\cdot \Phi ( \mathbf{x_i}) - b) \ge 1-\xi_i\\
s.t \quad \forall i = 1, \dots, n
\label{eq_svm_soft_opti_skewed}
\end{split}
\end{equation}

The objective function of the standard SVM classifier shown in (\ref{eq_svm_soft_opti_skewed}) has two terms. The first part tries to maximize the margin while the second part attempts to minimize the penalty associated with the misclassification. The regularization parameter $C$, which is a constant, balances the trade-off between these two terms and can also be considered as the defining factor of the misclassification cost. As both the  positive and negative examples are assigned with the same misclassification  cost ($C$), the penalty term is minimized when the misclassification cost of the training samples become minimized. However, for an imbalance dataset with a higher number of negative examples  there could be more negative examples even near the class boundary where the ideal hyperplane is passing  through.  This influences the optimization problem, because in order to minimize the total misclassification cost, the separating hyperplane could shift towards the minority class. This undesirable shift can cause more false negative examples by impacting the performances of the minority class. However, the total misclassification cost and the  total error would remain low due to the higher number of true negative examples. In cases like affordance detection, where class imbalance is extreme, the SVM could easily produce a skewed hyperplane which would classify all examples as negative. 

\section{Imbalance Leaning Methods for SVMs}

As discussed in the previous section behaviour of the SVM classifier with imbalanced data is undesirable. However in many domains training data sets are highly skewed and therefore researchers have tried a variety of techniques to learn SVM models with imbalanced data. This section discusses a few widely used  methods and their applicability for learning affordances. 

The SVM classifier with imbalanced data can be  categorized into two main classes as data pre-processing methods and algorithmic modification methods.

\subsection{Data pre-processing methods}

In data pre-processing methods classifiers are trained on subsets of the original dataset and try to minimize the effects of class imbalance by balancing the training example in the subset. Most of these methods can be used in many classification algorithms that are affected by class imbalance and are not limited only to the  SVM algorithm. This section introduces  two such popular data preprocessing methods called re-sampling method and ensemble learning method.

\subsubsection{Re-sampling methods}

Re-sampling methods attempt to correct the data imbalance by balancing the number of training samples belonging to  each class and training the SVM on the modified balanced dataset. Many researchers  have applied re-sampling methods to train SVMs with imbalanced datasets in different domains \cite{wu2003adaptive, akbani2004applying, chawla2002smote, dehmeshki2004classification}. These methods can be categorized as  random  under/oversampling, focus under/oversampling and data generation methods like SMOTE \cite{chawla2002smote}. 

The basic SVM classifier presented in the previous section is based on random under sampling. In random under sampling, examples from the majority class are randomly selected and removed from the original dataset until the number of examples from both classes are reasonably balanced. Consequently, under sampling readily gives  a simple method for adjusting the balance of the original dataset. However, as shown in   the previous section  under sampling throws away potentially informative examples from the majority class, and it could affect the decision boundary of the SVM adversely resulting a higher number of false positives.

While under sampling removes examples from the original dataset, random oversampling appends examples of the minority class to the dataset by creating multiple copies of examples from the minority class. Although this method balances the class distributions, it also introduces its own set of problematic consequences. Firstly, adding more examples increases the size of the dataset and learning with an extremely imbalanced dataset would not be feasible. On the other hand, randomly added new examples would not even affect the decision boundary if most of the majority class examples reside near the decision boundary. More specifically, random oversampling could lead to  over-fitting since oversampling simply appends replicated data to the original dataset. This will cause the decision rule to become too specific and could  worsen the classification performances on unseen testing data even for a classifier with high training accuracy \cite{he2009learning}.

In contrast to random sampling, focused sampling methods try to select the most informative data points around the decision boundary and use them to balance the two classes. In  \cite{batuwita2010efficient} researchers  present an efficient focused sampling method specifically  for SVMs. First,  the separating hyperplane is found by training an SVM model on the original imbalanced dataset and using it to select the most informative examples lying around the decision boundary. Then these selected examples are used to balance the dataset and  a new SVM model is trained on this new dataset. This method is more suited for the affordance learning because all of the negative examples are not required for the SVM learner  and therefore the final classifier can be trained on  more informative examples of the original dataset. 

\subsection{Algorithmic Modifications}

The re-sampling methods discussed previously modify the original imbalanced datasets and apply them into the general SVM framework without modifying the SVM algorithm. However, SVM algorithm can be modified in order to  handle class imbalance internally. 

zSVM is such a widely used SVM imbalance training method proposed in \cite{imam2006z}. In this method first the standard SVM classifier is learnt on the original dataset. Then the bias of the separating hyperplane towards the minority class (positive class) is removed by artificially shifting the decision boundary toward the minority class.

The main reason for the inability of the soft margin SVM to learn the  separating hyperplane accurately when the dataset is imbalanced would be that it assigns equal cost values for both the minority and majority class misclassification in the penalty term. This would create a biased model with a  hyperplane that is  skewed  towards the minority class. The different cost model proposed in \cite{joachims1999transductive} has been designed to mitigate this issue by assigning different penalty terms for  two classes. By assuming that the positive class is the minority class, misclassification cost of $C^+$ can be assigned for the positive class and $C^-$ cost can be assigned for the negative class in the objective function of the soft margin SVM optimization equation.

\subsection{Preliminary Test Result}

It is important to understand the effectiveness and limitations of the SVM algorithms discussed in the previous section for affordance learning with highly imbalanced datasets. Therefore a series of experiments were carried out with different class imbalances and recorded the average F1-score for each class imbalance ratio.

Table \ref{Table-imbalance_test} summarizes the results of the imbalance test and shows the highest recorded F1-score and corresponding class imbalance ratio. Focus re-sampling method, DCM-SVM method and zSVM method have recorded better F1-scores than the random sampling methods in extreme class imbalances. However,  it is hard to predict the behaviour of these algorithms in different class imbalances. The  observations obtained in the class imbalance test have also revealed the following implementation difficulties.

\begin{table}[]
\centering
\caption{Imbalance Test Result Summary}
\label{Table-imbalance_test}
\begin{tabular}{l|l|c|c|}
\cline{2-4}
                                                                                                                      & {\bf Affordance Type} & {\bf Highest F1-Score} & {\bf Imabalance Ratio} \\ \hline
\multicolumn{1}{|l|}{\multirow{4}{*}{{\bf Random Sampling}}}                                                          & Sitting-Relaxing      & 0.35                   & 1:8                   \\ \cline{2-4} 
\multicolumn{1}{|l|}{}                                                                                                & Standing-Working      & 0.19                   & 1:2 \\ \cline{2-4} 
\multicolumn{1}{|l|}{}                                                                                                & Sitting-Working       & 0.48                  & 1:32 \\ \cline{2-4} 
\multicolumn{1}{|l|}{}                                                                                                & \multicolumn{3}{l|}{}                                                   \\ \hline
\multicolumn{1}{|l|}{\multirow{4}{*}{{\bf Focus Sampling}}}                                                           & Sitting-Relaxing      & 0.35                   & 1:2 \\ \cline{2-4} 
\multicolumn{1}{|l|}{}                                                                                                & Standing-Working      &  \bf{0.23}             & 1:4 \\ 
\cline{2-4} 
\multicolumn{1}{|l|}{}                                                                                                & Sitting-Working       & 0.38                   & 1:64 \\ \cline{2-4} 
\multicolumn{1}{|l|}{}                                                                                                & \multicolumn{3}{l|}{}                                                   \\ \hline
\multicolumn{1}{|l|}{\multirow{4}{*}{{\bf zSVM method}}}                                                              & Sitting-Relaxing      & \bf{ 0.36 }            & 1:8
 \\ \cline{2-4} 
\multicolumn{1}{|l|}{}                                                                                                & Standing- Working     & 0.15                   & 1:2 \\ \cline{2-4} 
\multicolumn{1}{|l|}{}                                                                                                & Sitting-Working       & 0.37                   & 1:2 \\ \cline{2-4} 
\multicolumn{1}{|l|}{}                                                                                                & \multicolumn{3}{l|}{}                                                   \\ \hline
\multicolumn{1}{|l|}{\multirow{3}{*}{{\bf \begin{tabular}[c]{@{}l@{}}Different Cost Model\\ (DCM-SVM)\end{tabular}}}} & Sitting-Relaxing      & 0.34                   & 1:4                  \\ \cline{2-4} 
\multicolumn{1}{|l|}{}                                                                                                & Standing-Working      & 0.14                   & 1:2                  \\ \cline{2-4} 
\multicolumn{1}{|l|}{}                                                                                                & Sitting-Working       & \bf {0.52}             & 1:1024                 \\ \hline
\end{tabular}
\end{table}

The first difficulty is, how to select the best SVM algorithm for inferring affordances in a new room. One option is to select the SVM algorithm that outputs the best  F1-score on a validation set. According to the results of the imbalance test, the three affordances have recorded their highest F1-score in three different algorithms.  If such a method was chosen then three different SVM models that are trained on three different algorithms would be selected for inferring affordances. However,  it is difficult to generalize such an approach to a large set of affordance types. 

The second difficulty is to incorporate all training data into the training set. This is found to be difficult because of the following reasons. Firstly, all the algorithms tested in the imbalance test have recorded low performances when the class imbalances become extreme although some of the algorithms have shown  some resistance to moderate class imbalances. Secondly, training becomes computationally difficult with a large set of examples as all examples are considered for optimization. This was  evident in the imbalanced test as the training beyond 1:2048 imbalanced ratio was difficult due to computer memory limitation (The test was done on a High Performance Cluster Computer with 16 GB memory).

On the other hand, except in the  DCM-SVM method, class imbalance is not considered in the SVM optimization problem. All these algorithms are first trained on the original imabalanced dataset and then the  separating hyperplane is shifted artificially to reduce the effect of the class imbalance. In the case of DCM-SVM, the method of selecting the cost model is rather arbitrary and  therefore the calculated hyperplane could be sub-optimum.

Because of these reasons SVM based algorithms discussed in this section are not satisfactory enough for learning affordance. Particularly, their performances on the 'standing-working' accordance is comparably 
low. The next section addresses these shortcomings and formulates the affordance learning as structured output learning which can be solved using the Structured output SVM (S-SVM) algorithm.

\section{Structured SVM for Learning Affordance Map}
The previous section discussed  SVM learning in imbalanced datasets and its inability to produce acceptable results when the class imbalance become severe. This is particularly true for affordance learning where even in a single image class imbalance is high. Therefore, when more images are added for training the class imbalance becomes extreme. As shown in the previous section, even some algorithms that showed good results in some other imbalanced datasets failed to provide acceptable results with affordance learning when the class imbalance become extreme. On the other hand, learning the conventional SVM in a large dataset with high dimensional data becomes practically infeasible.

However, similar class imbalances exist in some machine learning problems and the generalization of the conventional SVM for structured problems called Structured  SVM (S-SVM) has been applied successfully to learn discriminate  models in extreme class imbalances \cite{li2009landmark, tang2009svms}. Consider the example of binary text classification with the topic `machine learning' and `other topics'. As in many other text classification tasks, in the dataset of this problem there could be only less than 1$\%$ of documents with the topic `machine learning'. In these situations a performance measure like error-rate becomes meaningless as the classifier that classifies all documents as 'other' already has a great error-rate which is hard to beat in the optimization. To overcome this problem the information retrieval  community  has designed other performance measures like F1-score and precision/ recall which are meaningful even when the datasets are imbalanced. 

What does this mean for learning affordances in an imbalanced dataset? Instead of optimizing the error rate during training, which is what conventional SVMs and most other learning algorithms do, it seems like a natural choice to have the learning algorithm directly optimize on performance measures like the F1-score. This is why the binary classification needs to be defined as a structured  problem, since unlike error rate, F1-score is not a function of individual examples, but a function of a set of examples. For example, consider a  3D image with $n$ number of grid locations with labels $\mathbf{\bar{y}}_k=(y_1,...,y_n)$, $y_i \in \lbrace -1,+1\rbrace$ and a feature vector $\mathbf{\bar{x}}_k= (x_1,....,x_n)$,  $x_i \in \Re ^n$. Then for each predicted set of labels $ \bar{\mathbf{y}}_k \sp{\prime}$ there exist a F1-score $F(\mathbf{\bar{y}}_k,\bar{\mathbf{y}}_k \sp{\prime})$ with respect to the true labeling $\mathbf{\bar{y}}_k$ which can be optimized directly in training. This structured problem can be learned efficiently using Structured SVM (S-SVM). 

The S-SVMs inherit the attractive features of conventional SVM  namely a convex training problem, flexibility in the choice of loss function, and the opportunity to learn nonlinear rules via kernels. On the other hand it has the ability to build upon the underlying structure of data like most generative models (e.g., Markov Random Field). However, unlike generative models S-SVM does not assume feature independence and 
is a discriminative training model. This features make the S-SVM a suitable candidate for affordance learning.

The purpose of this section is to introduce the S-SVM algorithm for affordance learning. It first formulates the  affordance learning as a structured output problem by describing required algorithms. Then results of the quantitative and qualitative test are presented to evaluate its performances. As a contribution, this section formulates affordance mapping as a structured SVM problem and its performances are experimentally evaluated.

\section{Affordance Learning with SVM Optimized for  Performance measures}

Given a set of 3D pointcloud images $ \lbrace i_1,....,i_k\rbrace \subset I$ and their associated affordance-map  \textit{Y}, the goal is to learn mapping $ g:I \rightarrow Y$ which can be later used to automatically build an affordance-map in unseen pointclouds. 
 The apparent  solution to this problem is to define a hypotheses $h$ as a function that matches a single feature vector $x$ at skeleton location $X_L=\lbrace x,y,z,\theta\rbrace$ of each image  to a single label $y \in\lbrace-1, +1\rbrace $,

\begin{equation}
h:X \rightarrow Y
\label{h}
\end{equation}

Learning this mapping function is non-trivial due to  the challenges discussed in the previous section.  Therefore, the  following SVM algorithm that optimized on non-linear performance measures are proposed to learn and infer the affordances. This can be done by defining this learning problem as a multivariate prediction problem. 

Once  feature vectors $\mathbf{\bar{x}_k}= (\mathbf{x_1,....,x_n})$ for each  skeleton location $\mathbf{g}_j=(x_j,y_j,z_j,\theta_j)$  of the 3D pointcloud $k$ and its associated label $\bar{y}_k=(y_1,....,y_n)$ are known  the goal is to find a new function $\bar{h}$ that maps a tuple $\mathbf{\bar{x}_k} \in \bar{X}$ of $n$ feature vectors to a tuple of $\bar{y}_k  \in  \bar{Y} $ of $n$ labels

\begin{equation}
\bar{h}: \bar{X} \rightarrow \bar{Y}
\label{hbarl}
\end{equation}

where $\bar{X} = X \times ...\times X$ and $\bar{Y} \subseteq \lbrace -1, +1\rbrace ^{n}$ is the set of all possible label tuples. The following discriminant  function is used to implement the proposed multivariate mapping

\begin{equation}
\bar{h}_{w}(\bar{\mathbf{x}})= \arg\max_{\bar{y}^{'} \in \bar{Y}}\lbrace\mathbf{w}^T \Psi (\bar{\mathbf{x}_k},\bar{y}^{'})\rbrace
\label{argmax}
\end{equation}
 
 where $\mathbf{w}$ is the parameter vector and $\Psi$ is the feature function which defines the relationship between  features, $\mathbf{x}$ with the output $\bar{y}^{'}$. The above function $\bar{h}_{w}(\bar{\mathbf{x}})$ returns the set of labels $ \bar{y}^{'}=(y^{'}_1,......,y^{'}_n) $, which score high values according to the discriminant function, $\bar{h}_{w}(\bar{\mathbf{x}})$. Whether the $argmax$ in equation (\ref{argmax}) can be efficiently calculated solely depends on the structure of the feature function, $\Psi$. If the feature function is defined as a linear combination of labels and features as shown in (\ref{feature}) then (\ref{argmax}) can be calculated efficiently.

\begin{equation}
\Psi (\bar{\mathbf{x}}_k,\bar{y}_k)=\sum_{i=1}^{n}  y_{i} \mathbf{x}_i 
\label{feature}
\end{equation}

As the  feature function given in (\ref{feature}) is linearly decomposable over $\bar{y}^{'}$, the solution can be calculated by maximizing $\bar{y}^{'}$ element wise. This makes the computation very fast and efficient.

The major advantage of  multivariate rule $\hat{h}$ for SVM optimization is that it allows  the inclusion of a loss function $\Delta$ that  is   based on a set of examples ($F_1$ score, Precision $/$ Recall) rather than optimizing on a single example-based loss function like error rate.

To train the discriminant $\bar{h}_{w}$ function the following generalization of the support vector machine has been used. 

\subsection{Optimization Problem}

\begin{equation}
\begin{split}
 \min_{w,\xi}\dfrac{1}{2} \Vert w \Vert^{2} + C \sum_{i=1}^{m}\xi_k
 \\
\textrm{s.t.} \quad  \xi_k \geq 0, \forall k
\\
\mathbf{w}^{T} [\Psi (\bar{\mathbf{x}}_k,\bar{y}_k)-\Psi (\bar{\mathbf{x}}_k,{\bar{y} \sp{\prime}}]\geq \bigtriangleup ( \bar{y} \sp{\prime}, \bar{y}_{k}) - \xi_k ,\quad \forall k,\quad  \forall \bar{y} \sp{\prime}  \in \mathcal{Y} \setminus  \bar{y}_k
\label{optiequl}
\end{split}
\end{equation}

where $\bigtriangleup ( \bar{y} \sp{\prime}, \bar{y}_{k}) $ is a loss function that decreases during training as the the predicted tuple of outputs $ \bar{y} \sp{\prime}$ approaches the ground truth lable tuple, $ \bar{y}_{k} $.  This optimization is convex but in contrast to the  standard SVM the above optimization problem has an infeasibly large number of constrains (number of training examples into each possible tuple of  $\bar{y}_k  \in \mathcal{Y} $),  which makes this optimization intractable.  However, by using the  sparse approximation algorithm proposed by \cite{joachims2005support} the above optimization problem can be solved in a polynomial time for many types of multivariate loss functions. This will be detailed in the next section. The objective function that needs to be minimised (\ref{optiequl}), is a trade-off  between the model complexity, $ \parallel \mathbf{w}   \parallel$, and the hinge loss relaxation of training loss, $\sum{\xi_k}$. Here $C > 0$ is a constant that controls  this trade-off.

Therefore the cutting plane algorithm proposed by \cite{joachims2005support} has been used to iteratively train parameters of the discriminant function . It iteratively builds a  sub-set of working constrains,  $\mathcal{W}$ that is equivalent to the full set of constrains specified in (\ref{optiequl}) up to the precision $\varepsilon$. The algorithm first starts with an empty set of $\mathcal{W}$ and $\mathbf{w}=0$  iterates through the samples of the dataset. At each iteration,  the algorithm finds the most violated constraint (Line 5), $i.e$ the constraint related to the label  $\bar{y}^{'}$  that maximizes   $ \bigtriangleup ( \bar{y} \sp{\prime}, \bar{y}_{k})+  \mathbf{w}^T \Psi (\bar{\mathbf{x}_k},\bar{y}^{'}) $. If this constraint is violated by more than $\varepsilon$  (Line 6) then it is  added to the  the current working constraint set $\mathcal{W}$ and new $\mathbf{w}$ is calculated by solving the quadratic program over the new working constraint set,  $\mathcal{W}$  (Line 6). Finally the algorithm stops when no constraint of the optimization problem in (\ref{argmax}) is violated more than $\varepsilon$. Therefore the solution obtained through the algorithm fulfills all constraints up to the precision $\varepsilon$, and the norm of  $\mathbf{w}$ is not bigger than the norm of the exact solution. Interestingly the proposed cutting plane algorithm always terminates  in a polynomial number of iterations as  shown in  \cite{joachims2005support}. It has also been shown that the refined version of the algorithm terminated after adding most of $O(C\varepsilon^{-1})$ constraints that are independent of the size of the output space $\mathcal{\bar{Y}}$ and the number of training examples. This makes the proposed algorithm an attractive training solution to  learn the affordance-map. 

\subsection{Maximization Step at Inference}

Since complexity of the proposed structured SVM method largely  depends on efficient calculation of $argmax$s   in inference equation \ref{argmax} and cutting plane  training algorithm, it is important that these maximizations can be computed efficiently. At inference step we need to compute,

\begin{equation}
\bar{h}_{w}(\bar{\mathbf{x}_k})= \arg\max_{\bar{y}^{'} \in \bar{Y}}\lbrace\mathbf{w}^T \Psi (\bar{\mathbf{x}_k},\bar{y}^{'})\rbrace
\label{argmax2}
\end{equation}

By substituting feature function, (\ref{feature}) in (\ref{argmax2}) the following form of the inference equation can be obtained.

\begin{equation}
\bar{h}_{w}(\bar{\mathbf{x}_k})= \arg\max_{\bar{y}^{'} \in \bar{Y}}\lbrace\mathbf{w}^T \sum_{i=1}^{n}  y_{i}^{'} \mathbf{x}_i \rbrace
\label{argmax3}
\end{equation}

Since the feature function is linearly decomposable over single label, $y_i^{'}$  the above inference equation can be converted to a simple form as shown below.

\begin{equation}
\bar{h}_{w}(\bar{\mathbf{x}_k})= \arg\max_{\bar{y}^{'} \in \bar{Y}}\lbrace \sum_{i=1}^{n} \mathbf{w}^T  y_{i}^{'} \mathbf{x}_i \rbrace
\label{argmax4}
\end{equation}

The $argmax$ in equation (\ref{argmax4}) can be further decomposed into a single feature vector and single label as follows, which makes structured SVM equivalent to the conventional SVM prediction problem. 

\begin{equation}
 *y_i^{'}= \arg\max_{{y}_i^{'} \in \bar{y}^{'}}\lbrace  \mathbf{w}^T  y_{i}^{'} \mathbf{x}_i \rbrace ,\quad  \forall {y}_i^{'} \in \bar{y}^{'}
\label{argmax5}
\end{equation}

As each label is a binary value with $y_i^{'} =\lbrace +1,-1\rbrace$ computing prediction becomes very efficient with

\begin{equation}
\bar{h}_{w}(\bar{\mathbf{x}_k})= \arg\max_{\bar{y}^{'} \in \bar{Y}}\lbrace\mathbf{w}^T \Psi (\bar{\mathbf{x}_k},\bar{y}^{'})\rbrace= (sign(\mathbf{w}.\mathbf{x}_1),........,sign(\mathbf{w}.\mathbf{x}_n))
\end{equation}

Therefore, the linear feature function $\Psi$ makes predictions very fast  by considerably reducing  the size of the output space ($\bar{Y}$) from $2^n$ to $n$, where $n$ is the  number of different grid locations in the map.

\subsection{Maximization Step at Learning}

However, computing the loss-augmented argmax  that is required  for training is more complicated and depends on the type of loss function used. Therefore, maximization step at training needs a sophisticated algorithm that allows efficient training. 

The step that calculates the most violated constraint in  the cutting plane optimisation algorithm  needs to complete the following maximization step shown in equation (\ref{most_violated}).

\begin{equation}
\bar{y}^{'} \leftarrow  \arg\max_{\bar{y}^{'} \in \bar{Y}}\lbrace \bigtriangleup ( \bar{y} \sp{\prime}, \bar{y}_{k})+  \mathbf{w}^T \Psi (\bar{\mathbf{x}_k},\bar{y}^{'})\rbrace
\label{most_violated}
\end{equation}

If the loss function $\bigtriangleup ( \bar{y} \sp{\prime}, \bar{y}_{k})$ is linear in $\hat{y}^{'}$ (Eg : Hamming Loss) then the solution for (\ref{most_violated}) can be computed by maximizing $\bar{y}^{'}$ element wise. However, linear loss functions like, hamming loss cannot handle the label bias problem  associated with highly unbalanced classes as discussed in section 4. To overcome this problem, it seems like a natural choice to directly optimize for performance measures like $F_1$ score which is the harmonic mean of precision and recall.  This can be  done in the structural SVM framework by incorporating a loss function based on the contingency table at the maximization step.

\subsubsection{Loss Function Based on Contingency Table}
\label{sec_loss_function}
When the loss function becomes non-linear in $\bar{y}^{'}$, computation of the $argmax$ in Eq. (\ref{most_violated}) is  difficult because an exhaustive search over all possible $\bar(y)^{'}$ is infeasible. However, the computation of the $argmax$  can be categorized over all possible contingency tables so each sub-problem can be solved efficiently.
The contingency table for a binary classification  is shown in Table \ref{Tab_contigency}, where $a$ is the number of true positives, $b$ is the number of false positives, $c$ is the number of false negatives and $d$ is the number of true negatives.

\begin{table}[h]
\begin{center}
\begin{tabular}{|l|l|l|}
\hline   & y=1 & y=-1 \\ \hline
h(x)=1  & a   & b    \\ \hline
h(x)=-1 & c   & d    \\ \hline
\end{tabular}
\caption{Contingency table for Binary Classification}
\label{Tab_contigency}
\end{center}
\end{table}

It can be observed that there are only order  $O(n^2)$ of different  contingency tables for a binary classification problem with $n$ samples. Therefore, if the loss function $\bigtriangleup(a,b,c,d)$  is computed from the Table \ref{Tab_contigency} then the loss function  can  only have at most $O(n^2)$ different values. 

With the Loss function $\bigtriangleup(a,b,c,d)$, $argmax$ for calculating the most violated constraint can be redefined as  in (\ref{Eq_most_violated_abcd_loss}) .

\begin{equation}
\bar{y}^{'} \leftarrow  \arg\max_{\bar{y}^{'} \in \mathcal{\bar{Y}}}\lbrace \bigtriangleup ( a,b,c,d)+  \mathbf{w}^T \Psi (\bar{\mathbf{x}_k},\bar{y}^{'})\rbrace
\label{Eq_most_violated_abcd_loss}
\end{equation}

Although computation of $argmax$ in (\ref{Eq_most_violated_abcd_loss}) is exhaustive, it can be simplified by stratifying search space of $\mathcal{\bar{Y}}$ over different contingency tables. Algorithm \ref{contigency_al} details this process. For each contingency table $(a,b,c,d)$, it calculates $argmax$ over all $\mathcal{\bar{Y}}_{abcd}$ which is a subset of $\bar{y}^{'}$ that  correspond to the selected contingency table. 

\begin{equation}
\bar{y}_{abcd}= \arg\max_{\bar{y}^{'} \in \mathcal{\bar{Y}}_{abcd}}\lbrace\mathbf{w}^T \Psi (\bar{\mathbf{x}_k},\bar{y}^{'})\rbrace
\label{eq_argmax_abcd1}
\end{equation}

\begin{equation}
\bar{y}_{abcd}= \arg\max_{\bar{y}^{'} \in \mathcal{\bar{Y}}_{abcd}}\lbrace\mathbf{w}^T \sum_{i=1}^{n}  y_{i}^{'} \mathbf{x}_i \rbrace
\label{eq_argmax_abcd2}
\end{equation}

As in the inference function, the equation in (\ref{eq_argmax_abcd2}) is linear in $\bar{y}^{'}$. Therefore the solution can be obtained by maximizing $\bar{y}^{'}$ element wise. The maximum value of the object function for a particular contingency table is achieved when the $a$  positive examples with the largest value of $(\mathbf{w}^T \mathbf{x}_i)$ are classified as positive and the $d$ negative examples with the lowest value of $(\mathbf{w}^T \mathbf{x}_i)$ are classified as negative. Finally, overall $argmax$ can be found by  maximizing over each of these maxima plus their loss function value. 

\begin{algorithm}
 Input: $\mathbf{\bar{x}}_k=(\mathbf{x}_1,....,\mathbf{x}_n), \bar{y}=(y_1,.....y_n),\quad and \quad \bar{\mathcal{Y}}$ \;
$(i_1^{p},....,i_{\# pos}^{p}) \longleftarrow $  sort$\lbrace i:y_i=1\rbrace \quad $by $\mathbf{w}^T \mathbf{x}_i $ \;
$(i_1^{p},....,i_{\# neg}^{p}) \longleftarrow $  sort$\lbrace i:y_i=-1\rbrace \quad $by $\mathbf{w}^T \mathbf{x}_i $ \;

\For{$ a \in [0,....,\#pos]$}{
$c \quad \longleftarrow \quad \#pos-a $ \;
set $y^{'}_{i_1^{p}},.....,y^{'}_{i_a^{p}}$ to 1 AND set $y^{'}_{i_{a+1}^{p}},.....,y^{'}_{i_{\#pos}^{p}}$ to $-1$ \;

 \For{$ d \in [0,....,\#neg]$}{
$b \quad \longleftarrow \quad \#pos-d $ \;
set $y^{'}_{i_1^{n}},.....,y^{'}_{i_b^{n}}$ to 1 AND set $y^{'}_{i_{b+1}^{n}},.....,y^{'}_{i_{\#neg}^{n}}$ to $-1$ \;

$\upsilon \quad \longleftarrow \quad \bigtriangleup(a,b,c,d) + \mathbf{{w}}^T \sum_{i=1}^{n}  y_{i}^{'} \mathbf{x}_i $ \;
 
\If {$\upsilon$ is the largest so far}{

$\bar{y}^{*} \longleftarrow (y_1^{'},....,y_n^{'})$

 }

 }

  }

\Return{$\bar{y}^{*}$}

  \caption{for training structural SVMs (margin-rescaling)}
   \label{contigency_al}
\end{algorithm}

\subsection{Experiments}

This section explains implementation details of the affordance map building process using the structured SVM. 
The algorithms explained in previous sections can be easily incorporated into the publicly available  $SVM^{struct}$  \cite{tsochantaridis2004support} implementation which is a common framework to  predict multivariate or structured outputs. It needs the definitions of the following four functions.

The first is the feature function defined by (\ref{feature}).

\begin{equation}
\Psi (\bar{\mathbf{x}}_k,\bar{y}_k)
\label{feature_svm_struc}
\end{equation}

The second is the loss function,

\begin{equation}
\bigtriangleup ( \bar{y} \sp{\prime}, \bar{y}_{k})
\label{feature_svm_struc}
\end{equation}

which the  contingency table based loss function described in section. \ref{sec_loss_function}.

The third is a  function that calculates the most violated constraint.

\begin{equation}
\arg\max_{\bar{y}^{'} \in \bar{Y}}\lbrace \bigtriangleup ( \bar{y} \sp{\prime}, \bar{y}_{k})+  \mathbf{w}^T \Psi (\bar{\mathbf{x}_k},\bar{y}^{'})\rbrace 
\label{max_constraint_svm_struc}
\end{equation}

The algorithm \ref{contigency_al} is proposed to obtain  the solution for (\ref{max_constraint_svm_struc}).

Finally the $SVM^{struct}$ requires the inference function,

\begin{equation}
\arg\max_{\bar{y}^{'} \in \bar{Y}}\lbrace\mathbf{w}^T \Psi (\bar{\mathbf{x}_k},\bar{y}^{'})\rbrace
\label{inference_svm_strct}
\end{equation}

to predict the affordance map for a new image.

\section{Training}

Structured output SVM is a supervised training algorithm and therefore it needs to be trained before inferring the affordance-map in a new environment. First the feature set is calculated for each image of the training dataset, and cutting plane algorithm explained in \cite{joachims2005support} is used to calculate the weight vector, $\mathbf{w}$. Finally, the inference function given in  Eq. \ref{inference_svm_strct} is used to predict the affordance map in a new 3D image.

\begin{figure}
 \centering
\includegraphics[width=4.0in]{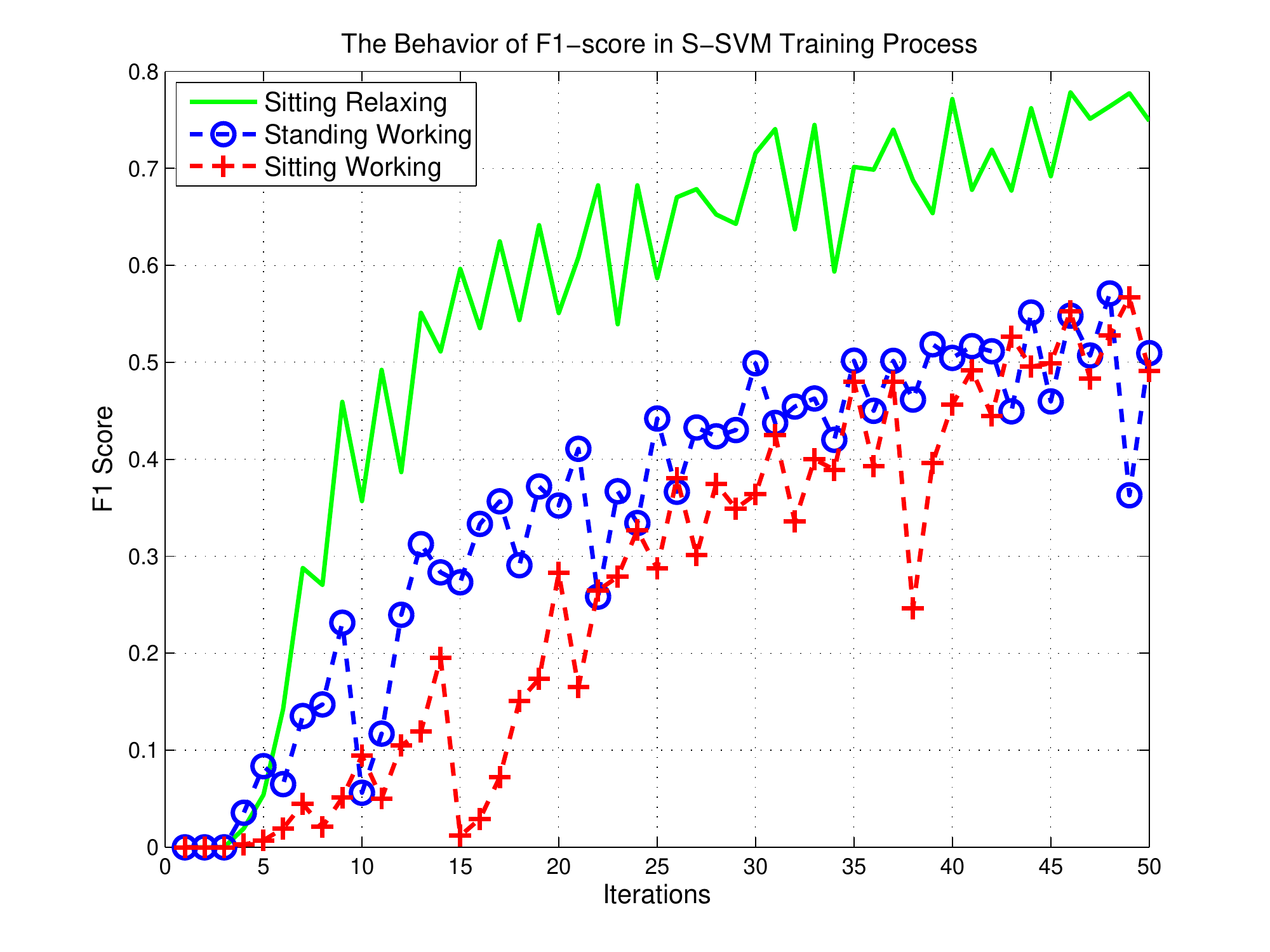}  
  \caption{The F1-score improves at each iteration of the S-SVM training process} 
  \label{fig_F1_score_bahavior}
\end{figure}

The main advantage of using structured output SVM training for affordance detection is,  parameter estimation can be directly optimized  on performance measures like F1-score, which are not affected by class imbalance problem.  Fig. \ref{fig_F1_score_bahavior} shows the behavior of the   F1-score  over each iteration of the training process.  The F1-score of sitting-relaxing affordance quickly improves and the F1-scores of the other two affordances have relatively slow improvements. This indicates the complexity of the training process associated with each affordance type. Fig. \ref{fig_F1_score_bahavior} also suggests that Sitting-Relaxing affordance is easy to learn when compared to other two affordances as its F1-score improves quickly.

As each iteration of the proposed structured output SVM algorithm improves the F1 score, the trained parameters could easily over-fit to the training data. This could degrade the performances on a previously unseen input image. However, over-fitting problem can be easily avoided by using a validation set. The optimum set of parameters for the structured SVM training is selected by recording the parameter set for each iteration of the training set and later testing them on a validation set. The parameter set that gives the best classification results on the validation set is used in inference.

\section{Experimental Results}

This section presents the experimental results of  the affordance mapping process based on the structured SVM algorithm. The k-fold cross validation is carried out to test and report the results. The classification results of the Structured output SVM  (S-SVM) are compared with the  methods discussed in the section 4, which previously  reported good performances in the imbalanced dataset.

In the following sections, experimental results are presented using qualitative and quantitative methods. The same framework used to report the results of the Naive Bayes and SVM classifier in section 3 is used for this purpose. 

\subsection{Qualitative Analysis}

An  qualitative method is used to compare the results between the S-SVM and other alternative  algorithms described in previous sections. It involves testing prediction results of the three affordances on a set of pre-selected 3D images. Both 2D affordance maps and 3D skeleton maps are used for this purpose.

\begin{figure*}
\centering
\subfigure[3D view of the living room]{%
\includegraphics[width=2.6in]{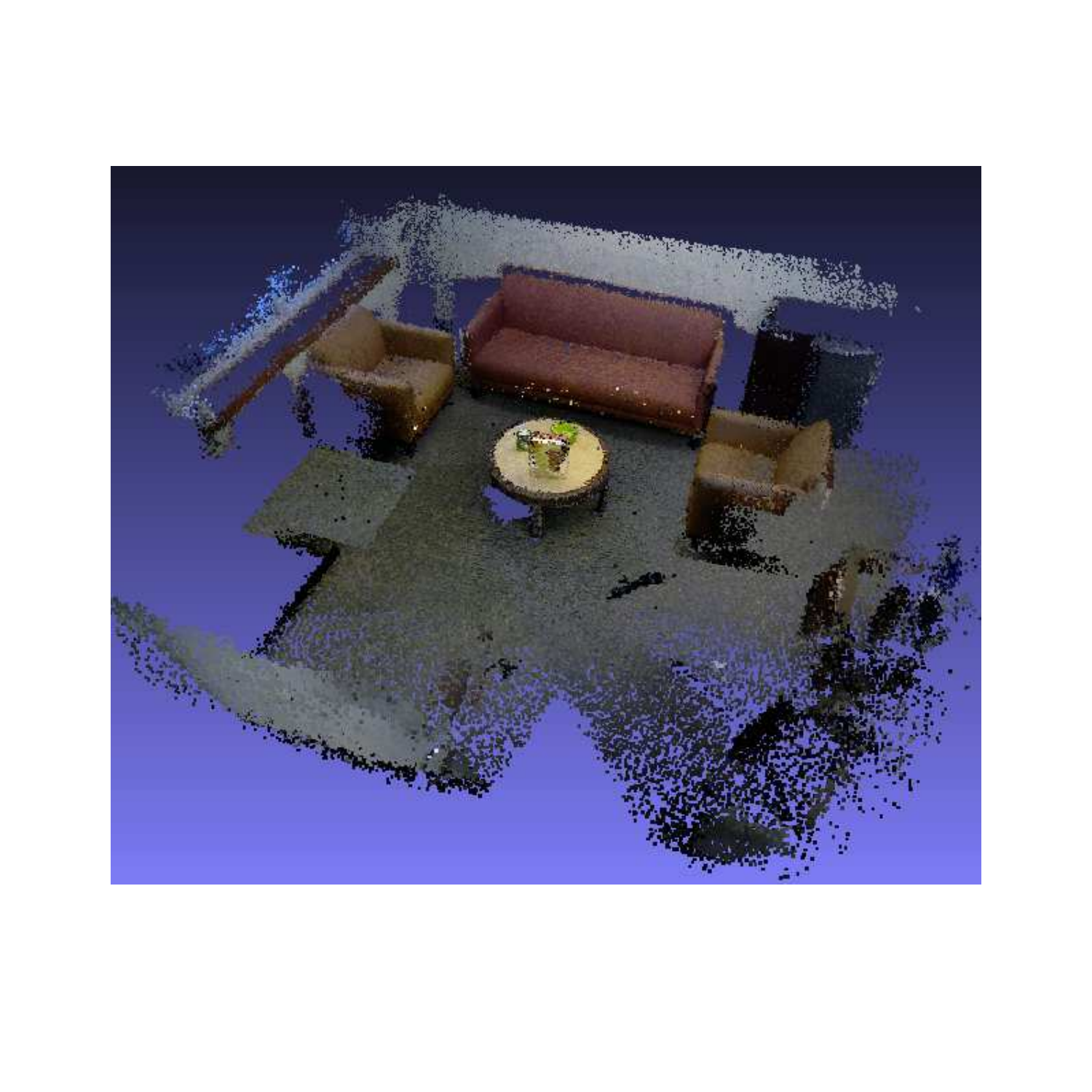}
\label{fig:room13d}}
\quad
\subfigure[Plan view with ground truth skeleton locations and their orientations]{%
\includegraphics[width=2.6in]{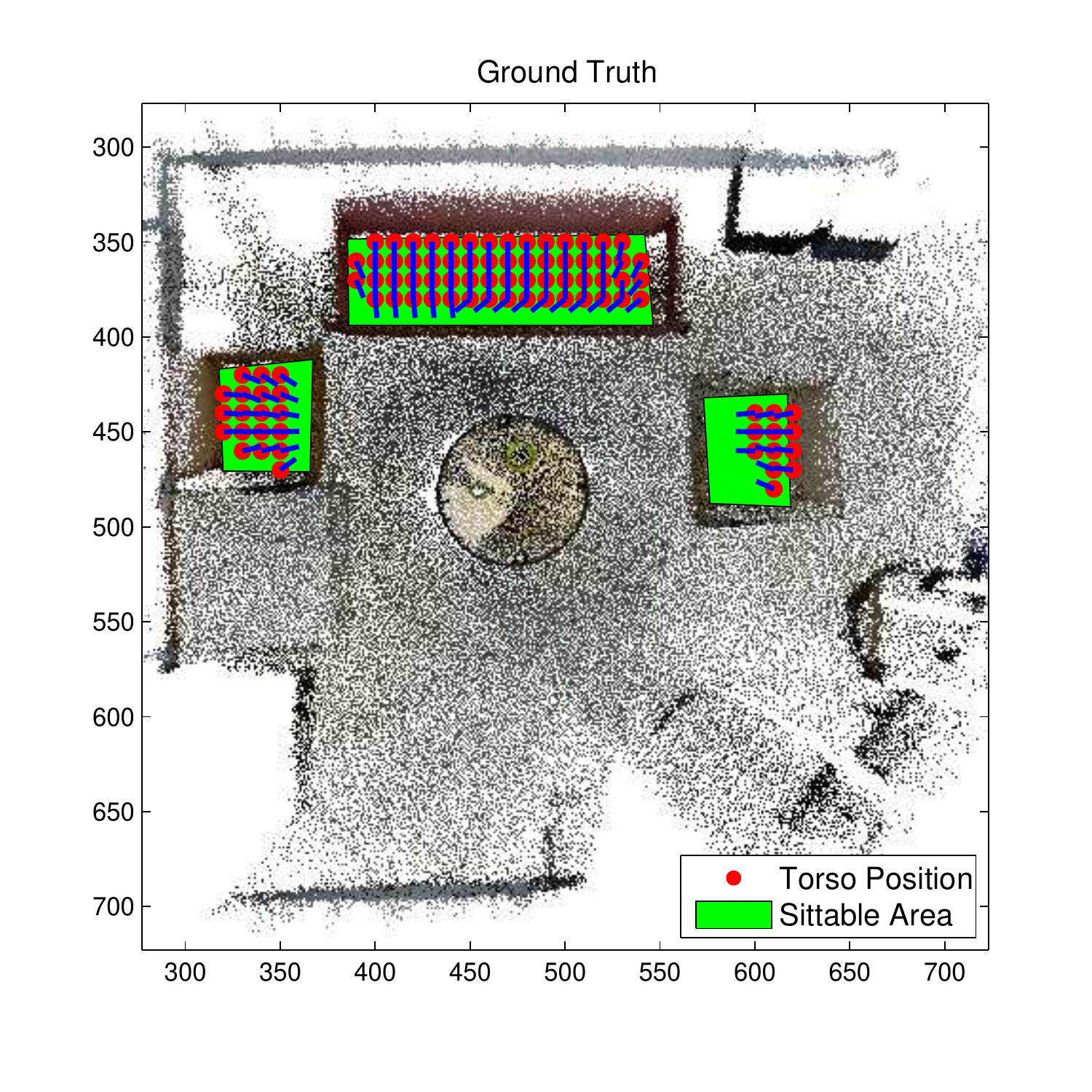}
\label{fig:room1_ground_ssvm}}

\centering
\subfigure[S-SVM Classifier]{%
\includegraphics[width=0.3\textwidth]{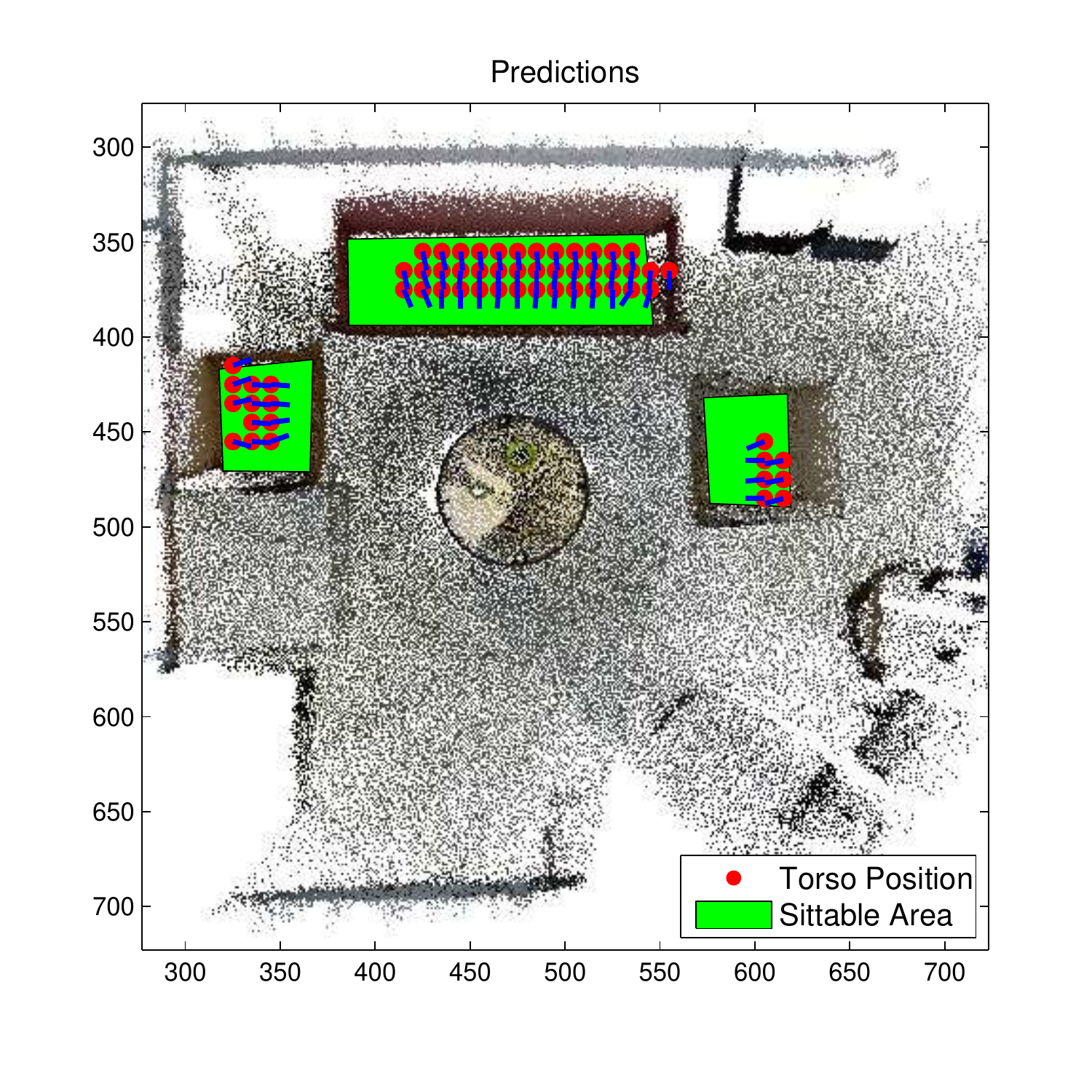}
\label{fig:prediction_relaxing_SSVM}}
\subfigure[S-SVM Classifier]{%
\includegraphics[width=0.3\textwidth]{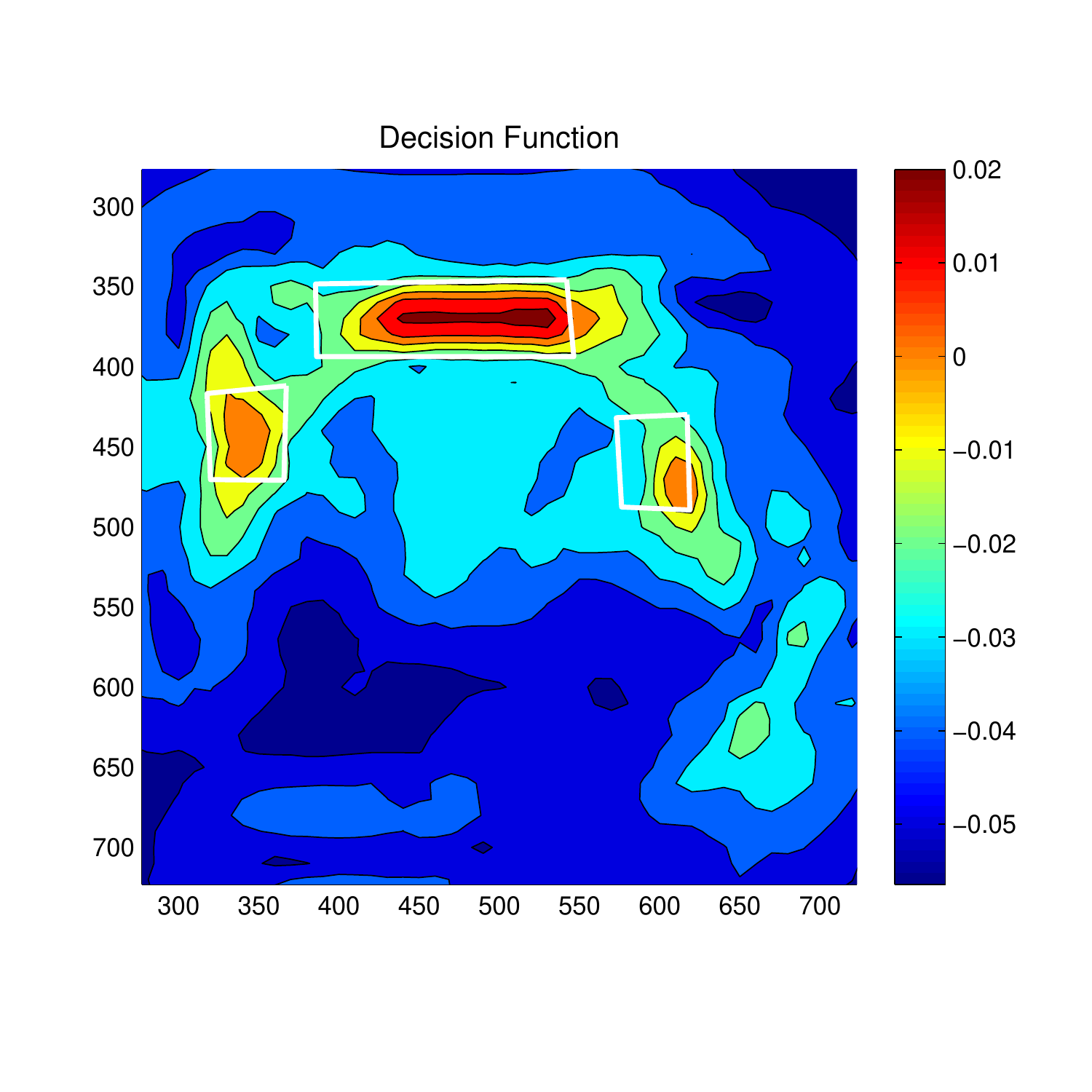}
\label{fig:decision_relaxing_SSVM}}
\subfigure[S-SVM Classifier]{%
\includegraphics[width=0.3\textwidth]{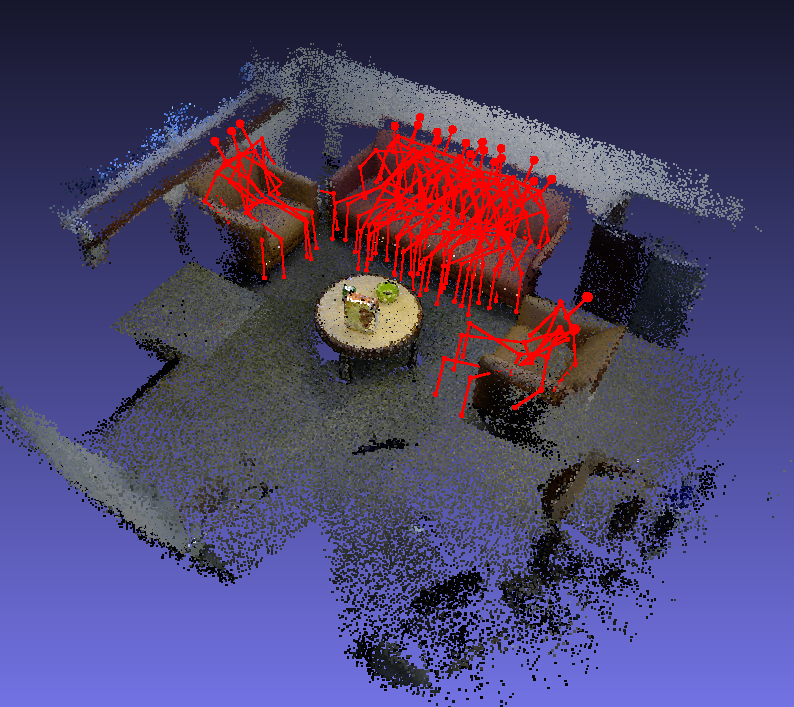}
\label{fig:skeleton_relaxing_SSVM}}
\\
\subfigure[zSVM Classifier]{%
\includegraphics[width=0.3\textwidth]{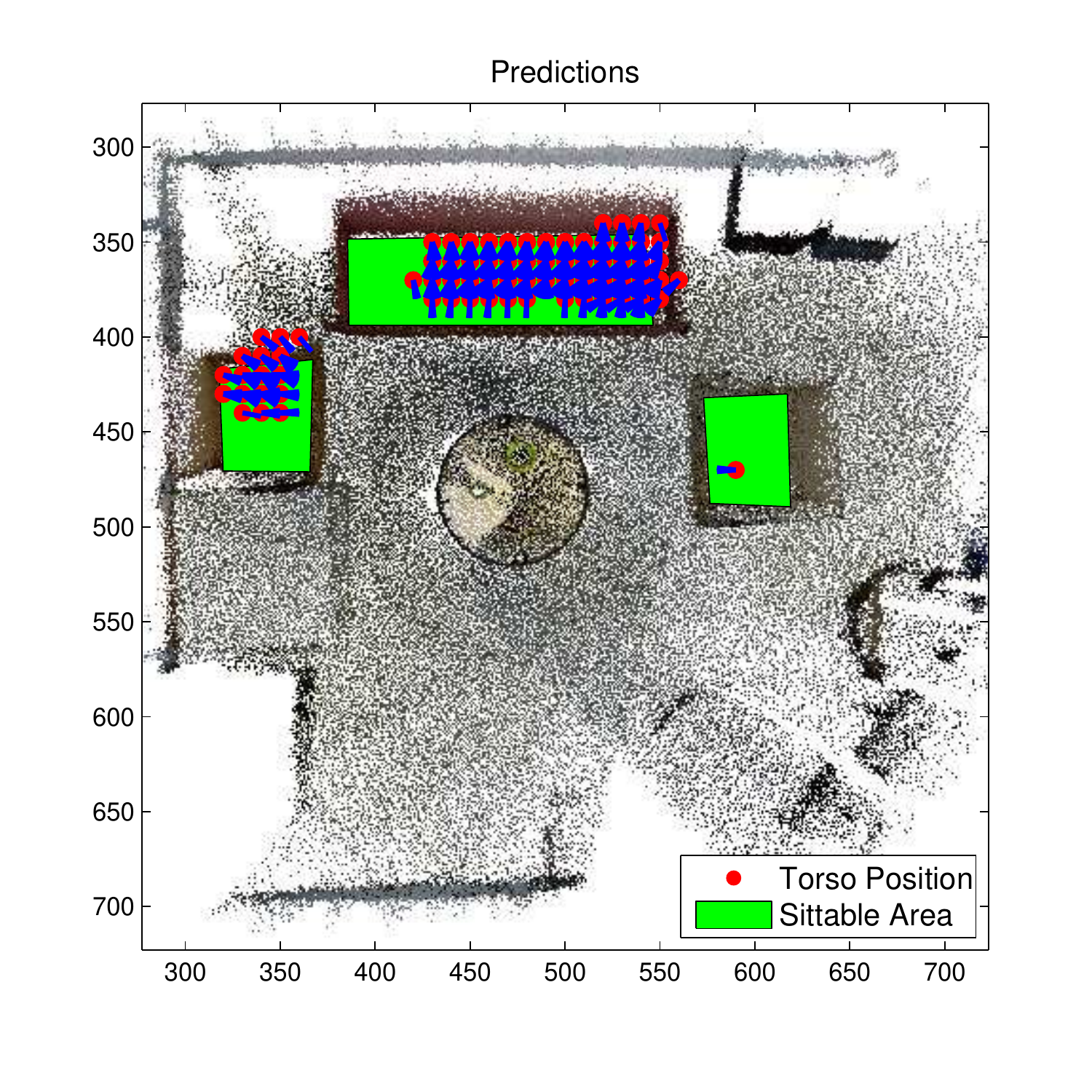}
\label{fig:prediction_relaxing_zSVM}}
\subfigure[zSVM Classifier]{%
\includegraphics[width=0.3\textwidth]{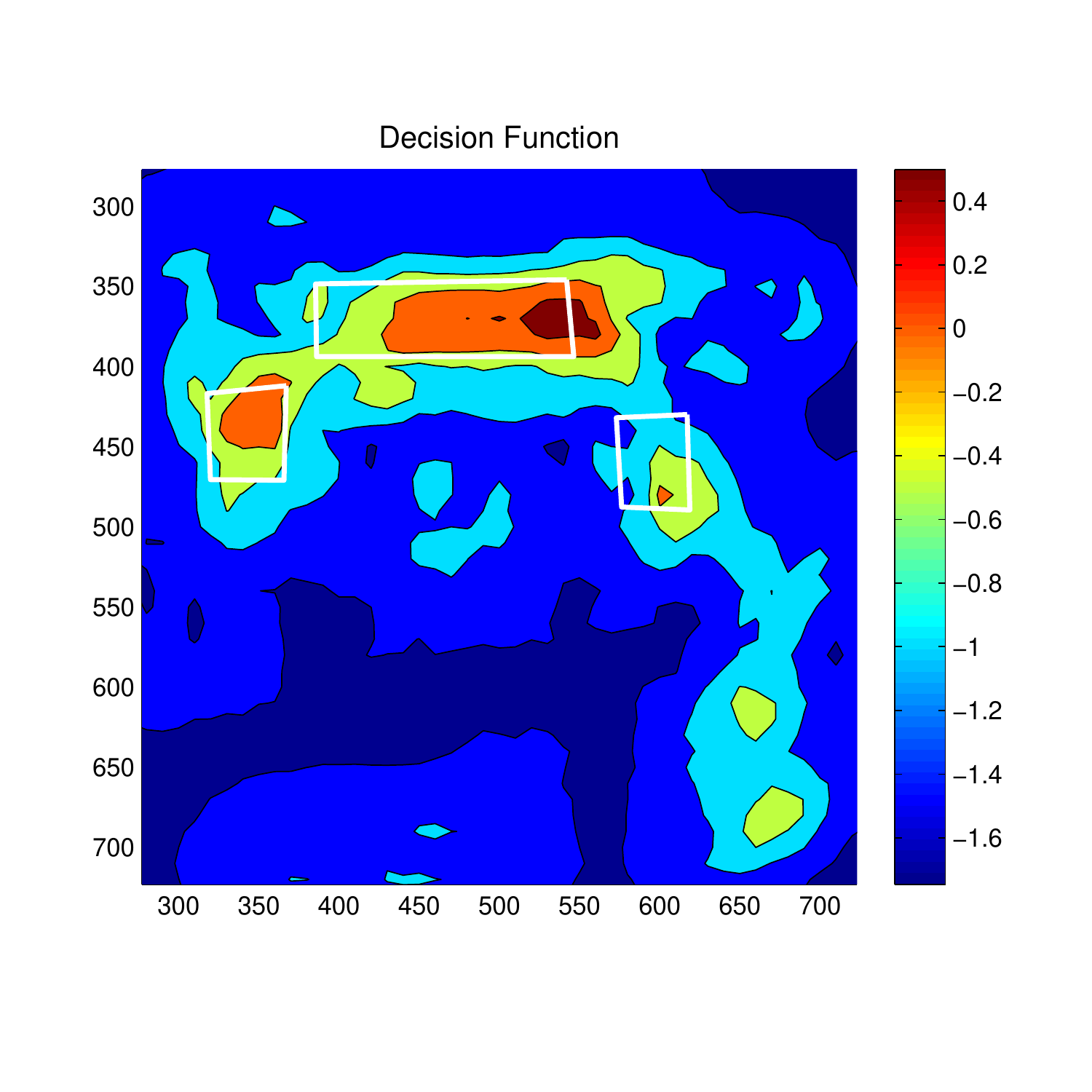}
\label{fig:decision_relaxing_zSVM}}
\subfigure[zSVM Classifier]{%
\includegraphics[width=0.3\textwidth]{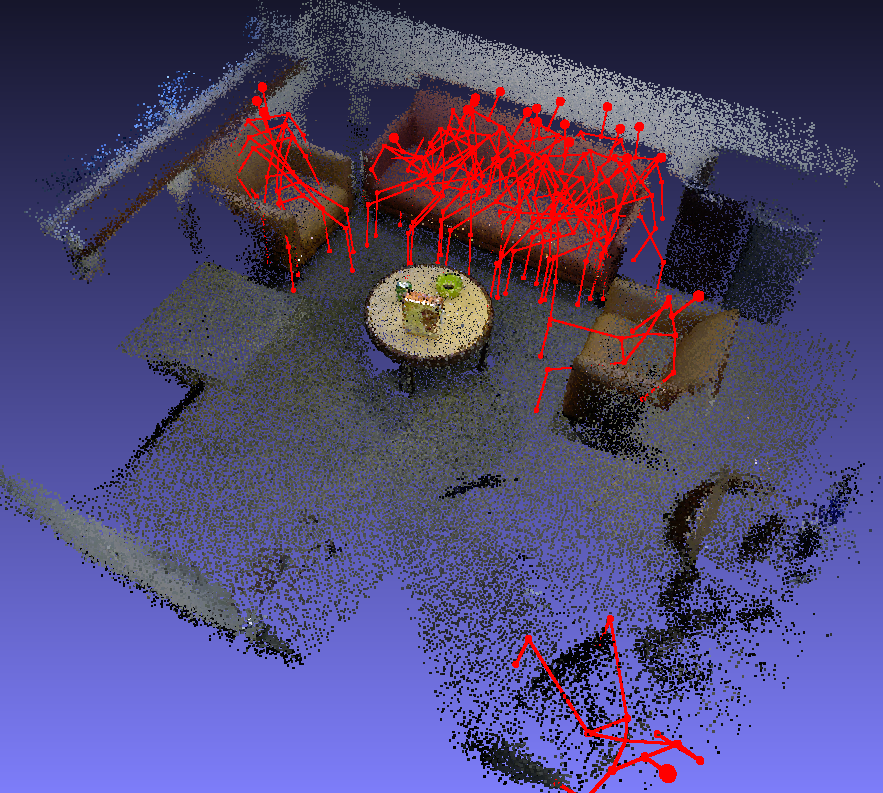}
\label{fig:skeleton_relaxing_zSVM}}
\caption{Classification results of the Sitting-Relaxing Affordance}
\end{figure*}

\subsubsection{Sitting-Relaxing Affordance}

 Fig. \ref{fig:room13d} shows the 3D view of a room and its ground truth skeleton locations of the room used to test the affordance `Sitting-relaxing'. It is the same room used in experiments explained in section 3. The predicted results of the S-SVM method are compared with the results of the zSVM method that previously recorded the highest F1-score for the Sitting-Relaxing affordance.

 Fig. \ref{fig:prediction_relaxing_SSVM} shows the prediction results of Structured Output SVM (S-SVM) and Fig. \ref{fig:prediction_relaxing_zSVM} shows the prediction results for the zSVM Model. Both these classifier have resulted in better predictions and the S-SVM has performed slightly better than the zSVM method. This is clearly  visible from the prediction results on the armchair towards the right-side of the room. On this armchair zSVM has only predicted one skeleton location while S-SVM has predicted several skeletons with a slightly better representation of the ground-truth.

\begin{figure*}

\centering
\subfigure[3D view of the office space]{%
\includegraphics[width=2.6in]{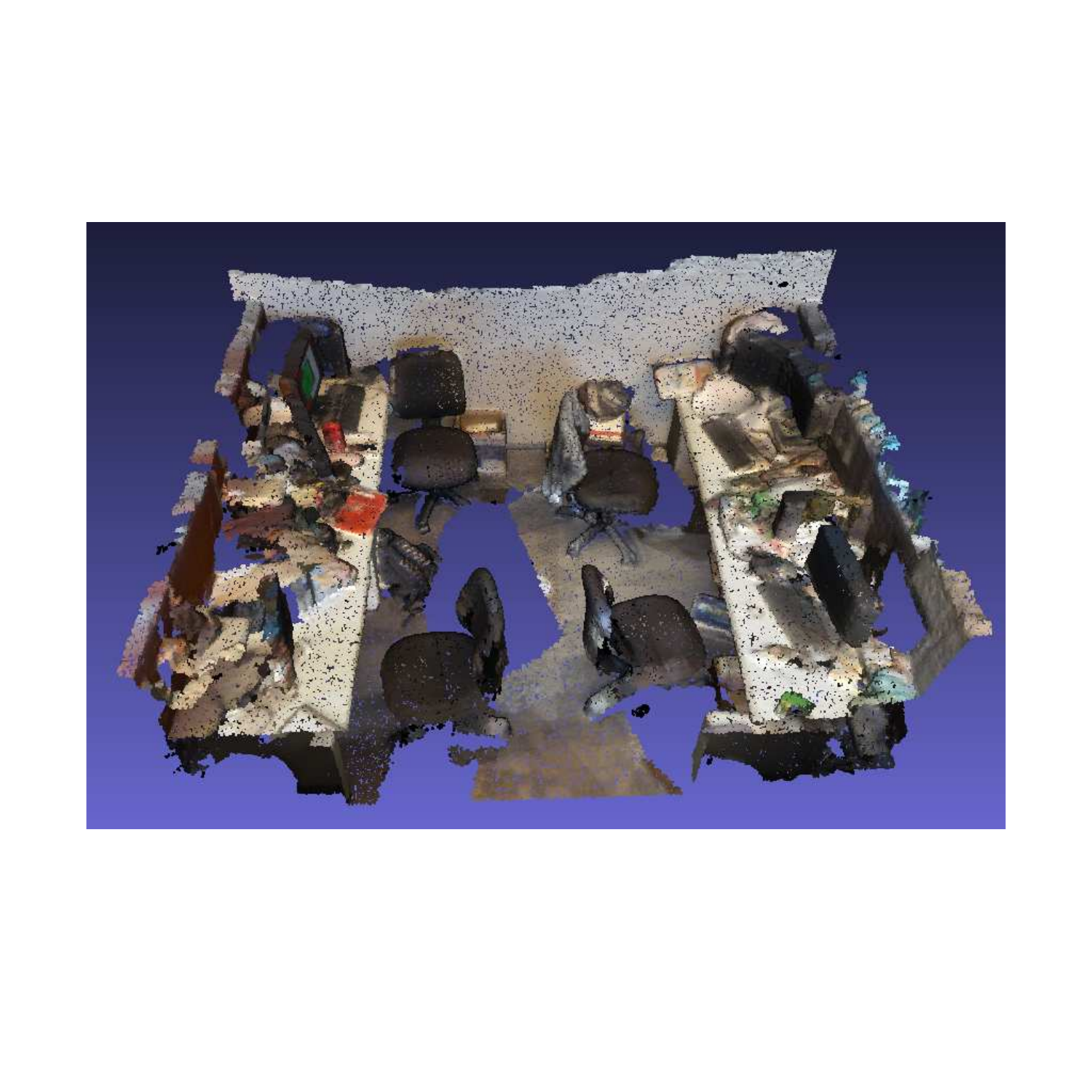}
\label{fig:room73d}}
\quad
\subfigure[Plan view with ground truth skeleton locations and their orientations]{%
\includegraphics[width=2.6in]{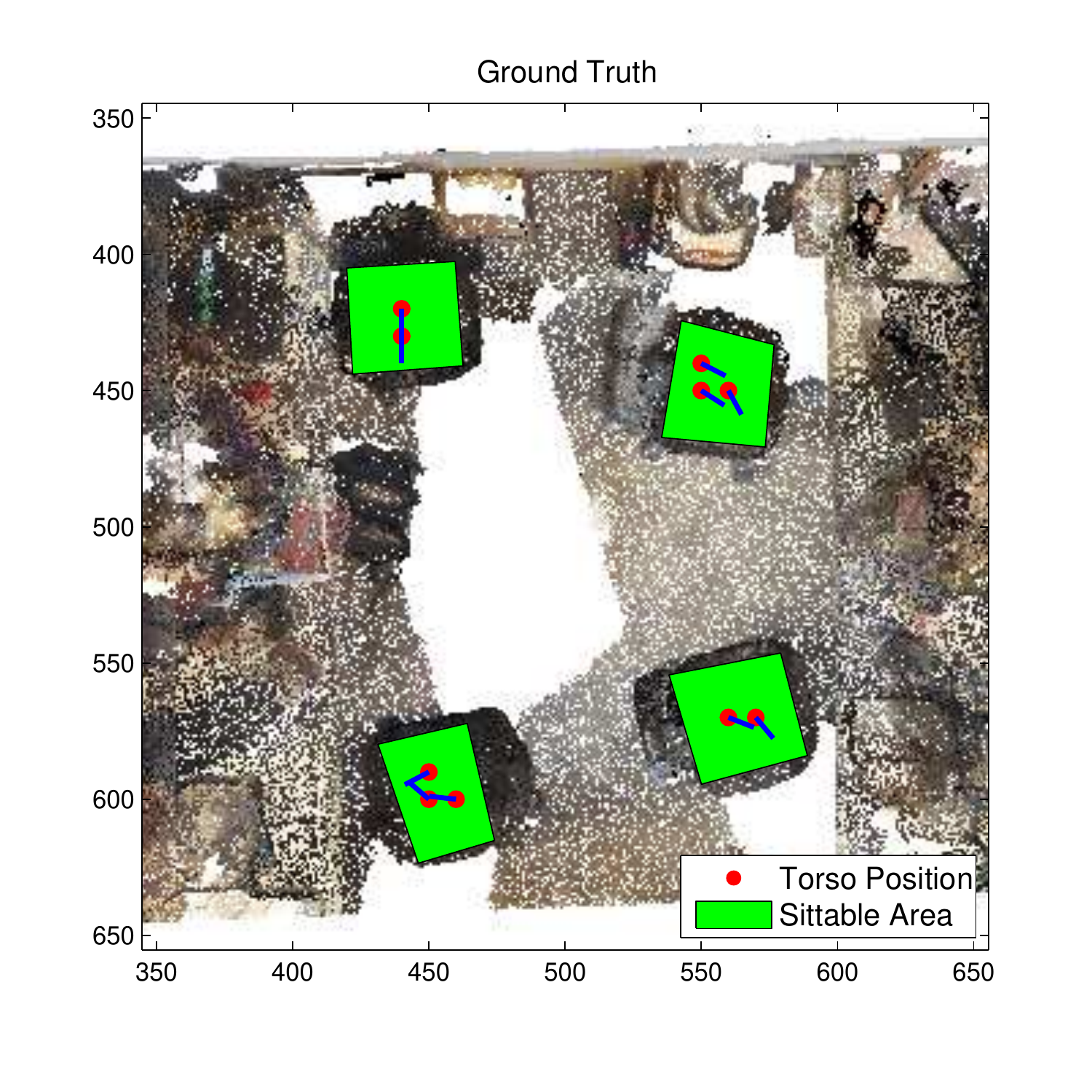}
\label{fig:room7_ground}}

\centering
\subfigure[S-SVM Classifier]{%
\includegraphics[width=0.3\textwidth]{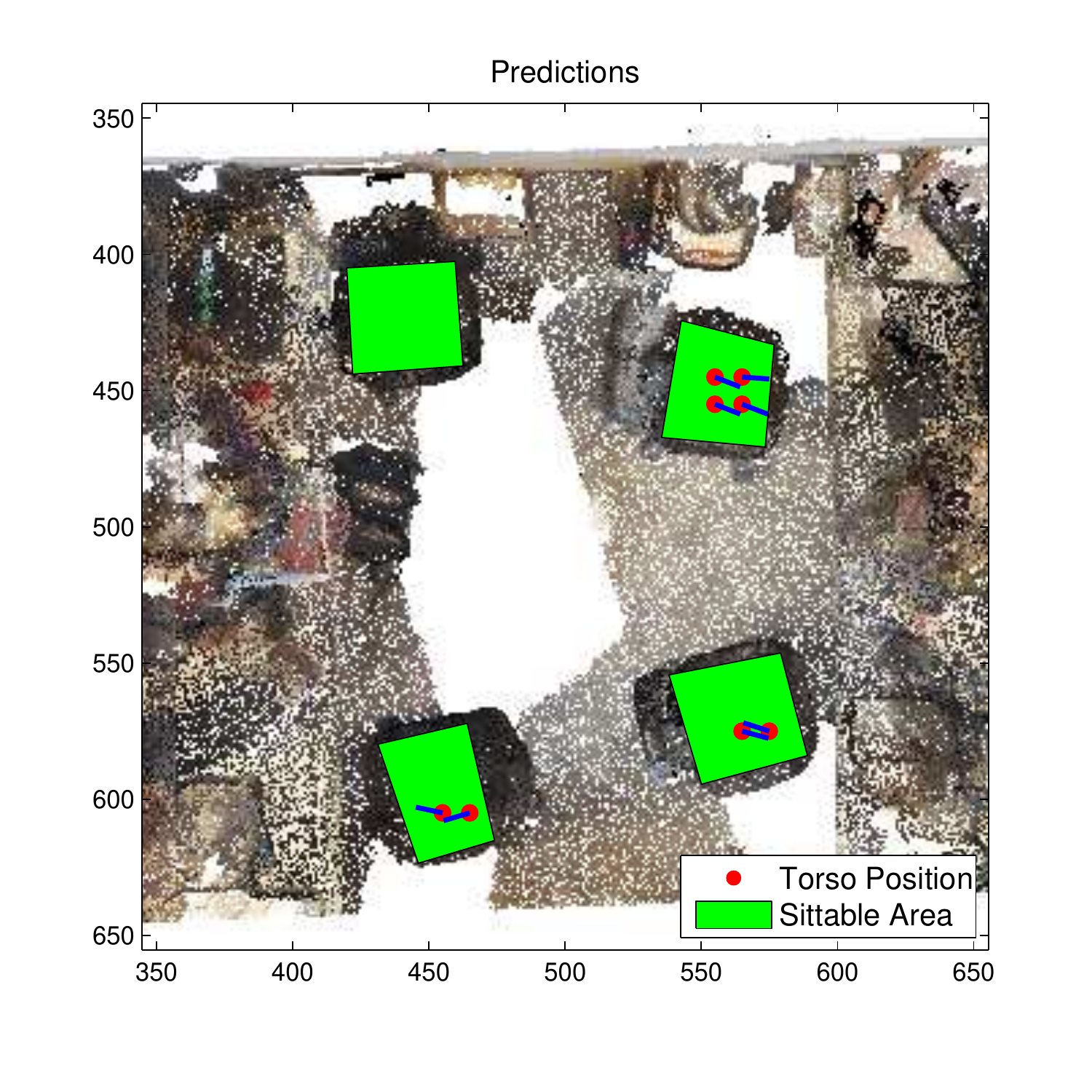}
\label{fig:prediction_working_SSVM}}
\subfigure[S-SVM Classifier]{%
\includegraphics[width=0.3\textwidth]{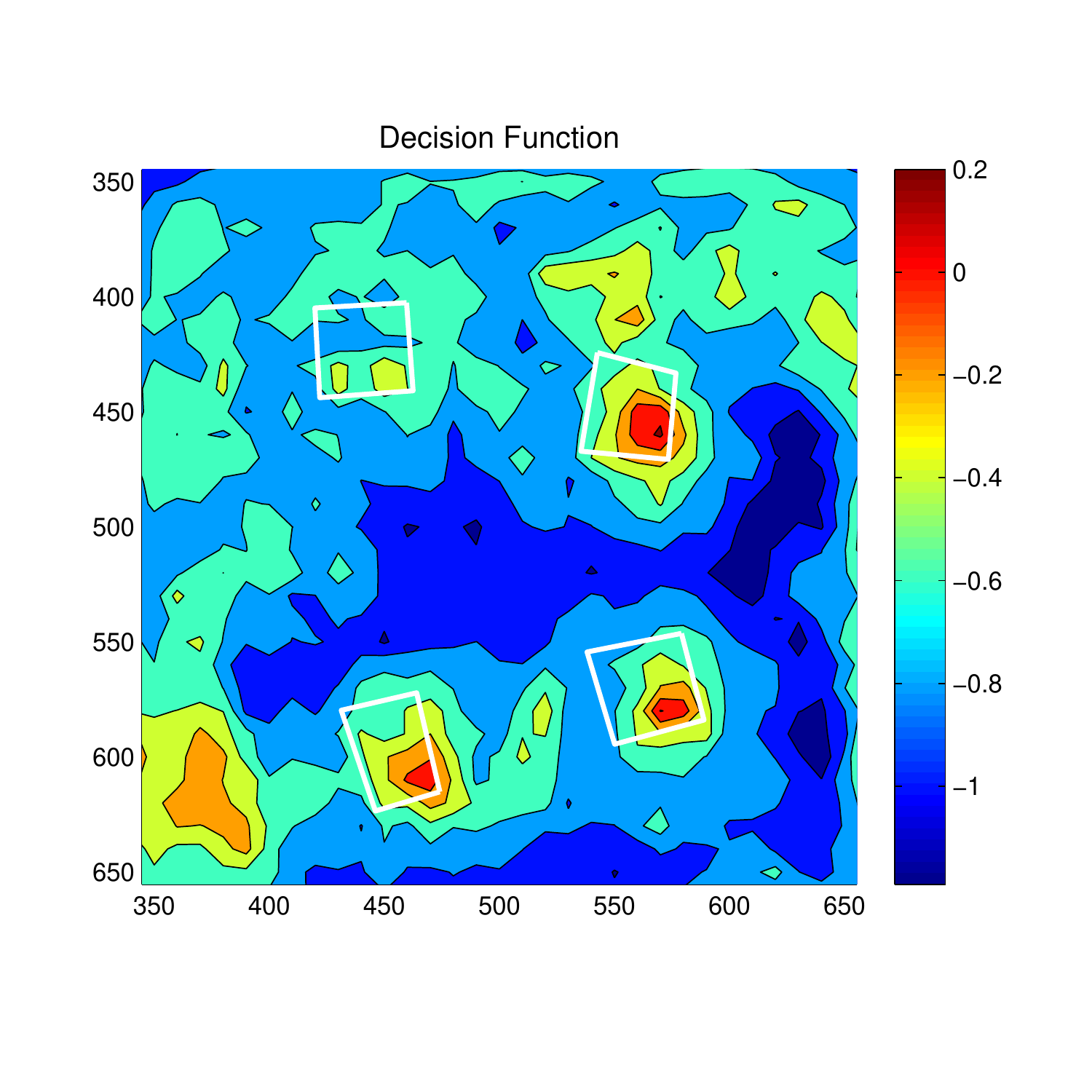}
\label{fig:decision_working_SSVM}}
\subfigure[S-SVM Classifier]{%
\includegraphics[width=0.3\textwidth]{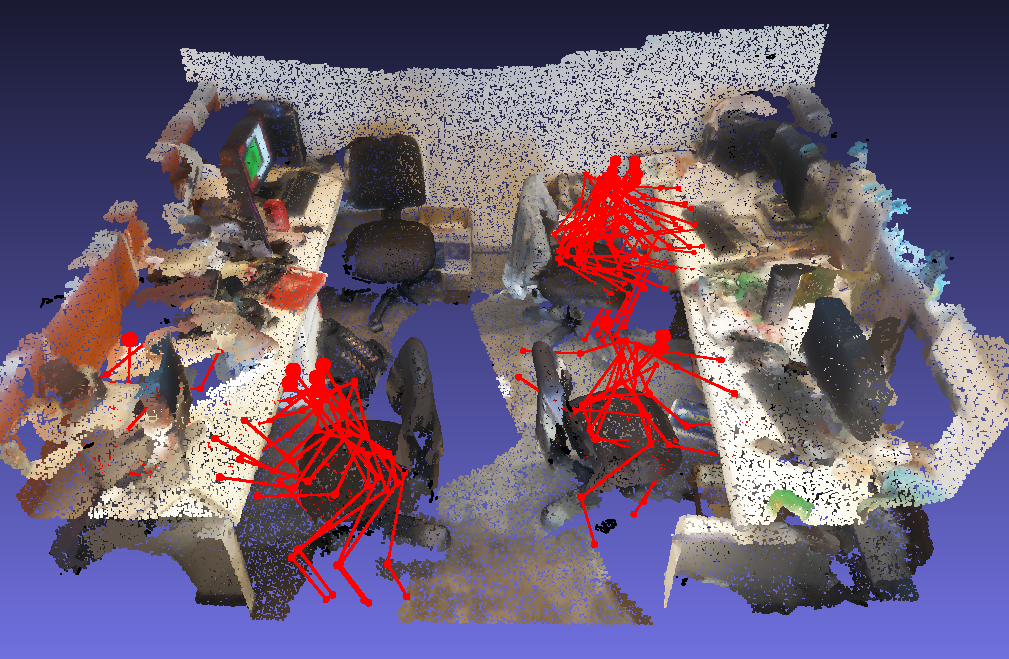}
\label{fig:skeleton_working_SSVM}}

\subfigure[DCM-SVM Classifier]{%
\includegraphics[width=0.3\textwidth]{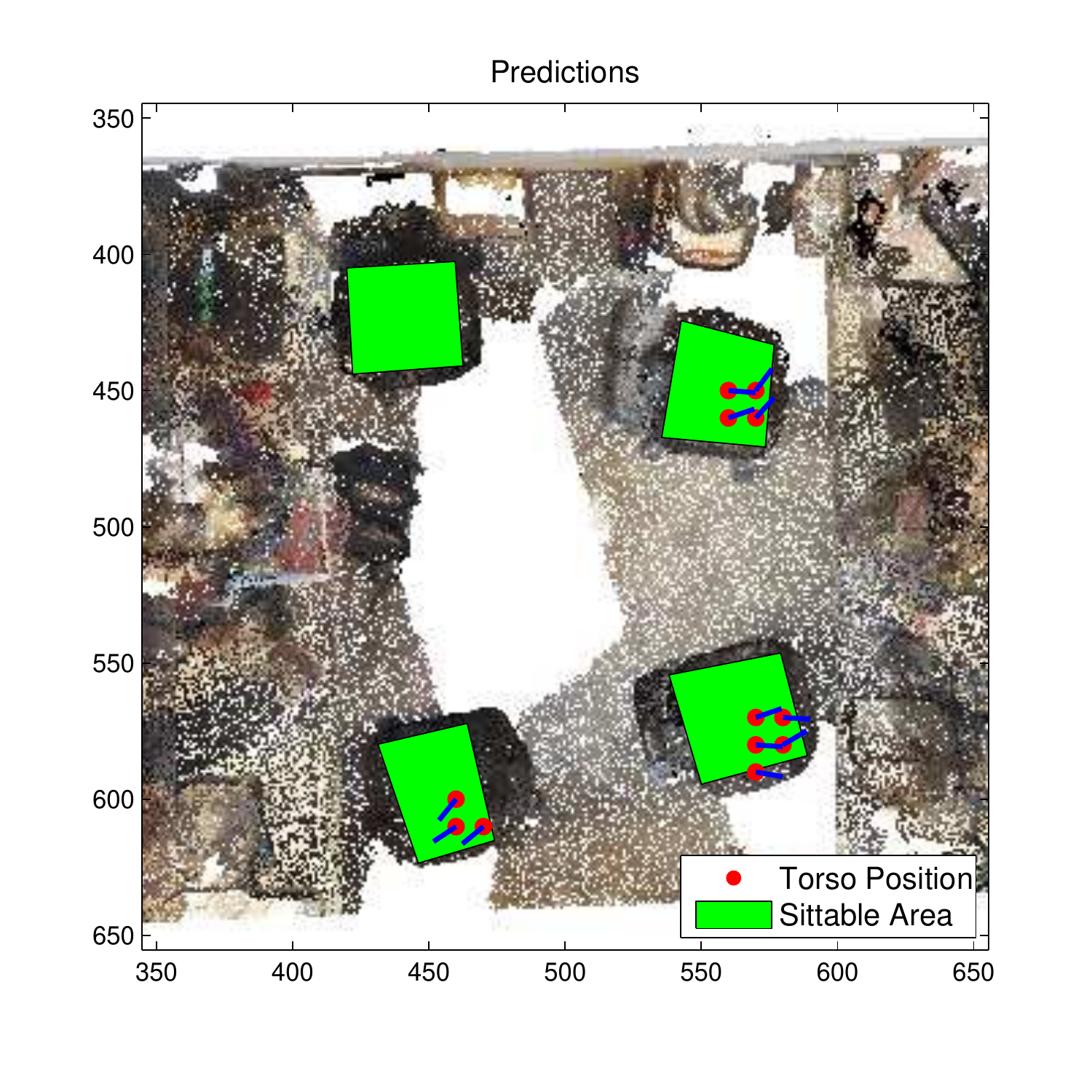}
\label{fig:prediction_working_DCM}}
\subfigure[DCM-SVM Classifier]{%
\includegraphics[width=0.3\textwidth]{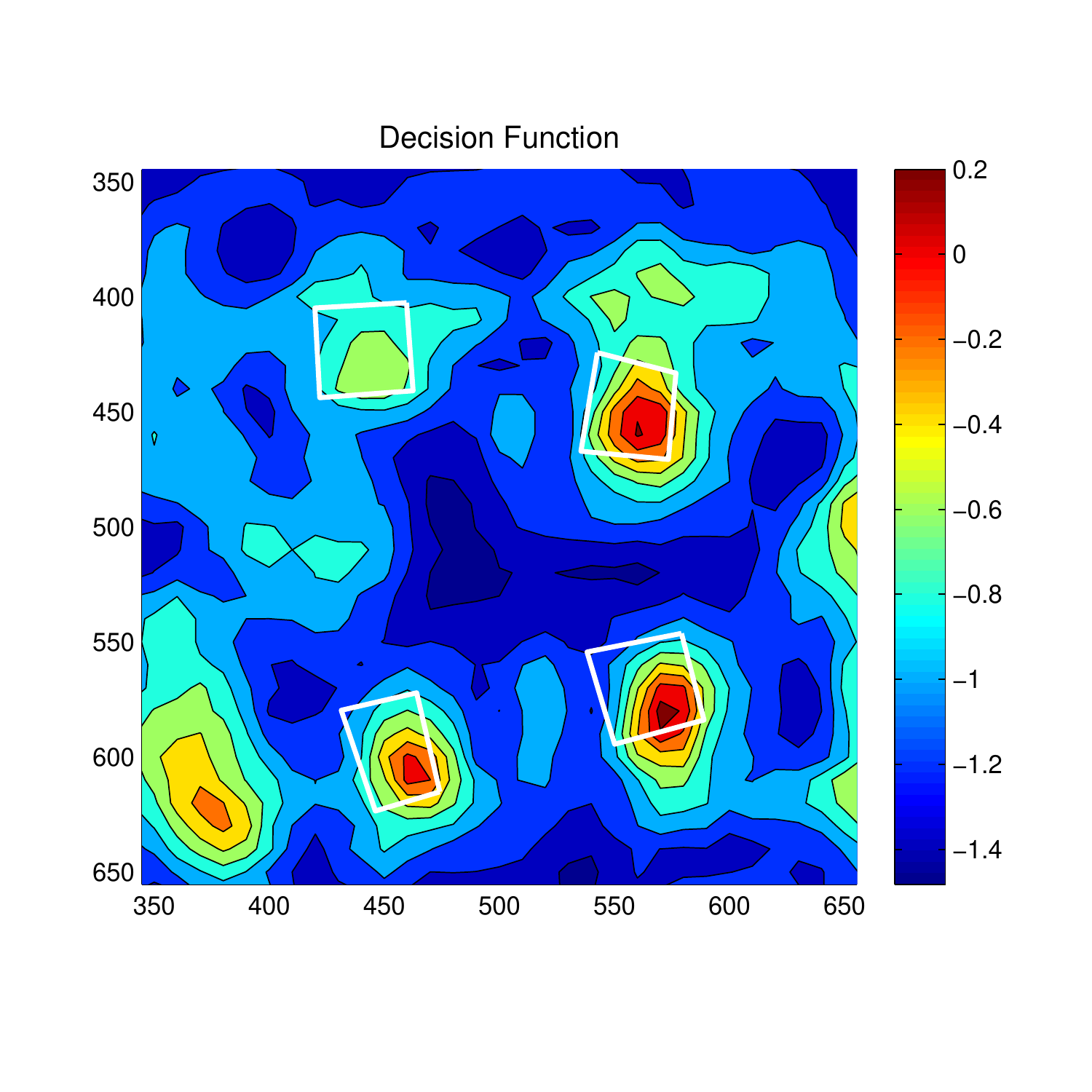}
\label{fig:decision_working_DCM}}
\subfigure[DCM-SVM  Classifier]{%
\includegraphics[width=0.3\textwidth]{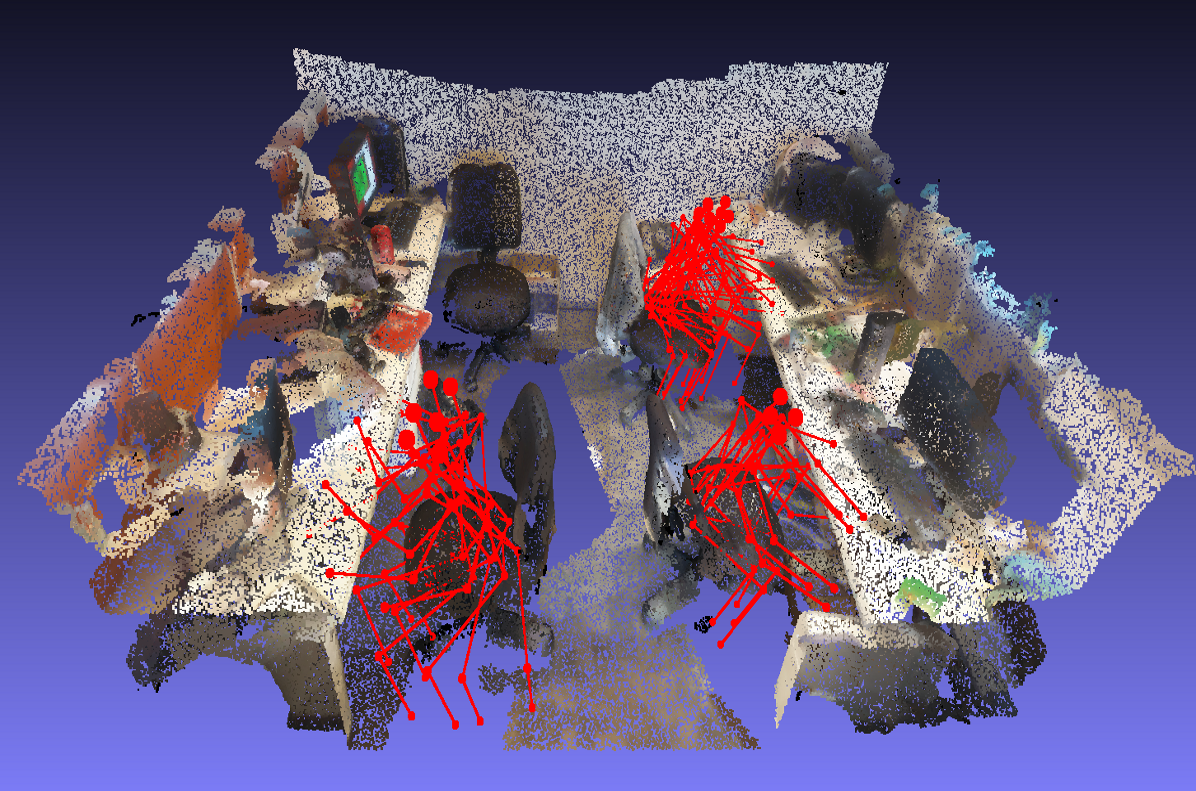}
\label{fig:skeleton_working_DCM}}
\caption{Classification results of the Sitting-Working Affordance}
\label{fig:prediction_working_SSVM_all}
\end{figure*}

Fig. \ref{fig:decision_relaxing_SSVM} and Fig. \ref{fig:decision_relaxing_zSVM}  show the decision function for the S-SVM classifier and the 
zSVM  classifier. It is clear from these results that the S-SVM decision function has better distribution of confidence values in positively predicted locations than the zSVM decision function. Specially, this can be observed near the long sofa towards the top-middle of the room. In this location, the S-SVM decision function has placed high confidence values throughout the sittable area of the sofa, whereas the zSVM decision function has shown a bias towards one of the corners of the sofa.

Fig. \ref{fig:skeleton_relaxing_SSVM} and \ref{fig:skeleton_relaxing_zSVM} show the skeleton map for the affordance type Sitting-Relaxing. It is clear that the S-SVM classifier has predicted 3D skeletons more accurately  than the z-SVM classifier. All the skeletons predicted by the S-SVM classifier are placed on the sofa set with correct orientation but the zSVM method has predicted few skeleton in locations other than the sofa. Further few of the skeletons predicted by zSVM on the sofa do not seems to be correctly oriented.

\subsubsection{Sitting-Working Affordance}

The 3D view and 2D plan view of the office space used to test the `sitting-working' affordance is shown in  Fig. \ref{fig:room73d}. It has four office chairs as shown with green rectangle areas. For the comparison of the results in the Sitting-Working affordance, the DCM-SVM method, which recorded the highest F1-score in the imbalance test, is used.

\begin{figure}[ht]

\end{figure}

The prediction results of the `sitting-working' affordance are shown in  Fig. \ref{fig:prediction_working_SSVM} and Fig. \ref{fig:prediction_working_DCM} . It is clear that both the S-SVM classifier and D-SVM classifier have predicted similar classification results. The two  classifier have accurately predicted `sitting-working' affordance on three chairs out of four chairs in the room but have failed to identify `sitting-working' affordance on the chair towards the top-left of the room. Also, one of skeletons predicted by the S-SVM classifier on the chair at the bottom-right of the room has oriented towards the back side of the chair.

\begin{figure*}
\centering
\subfigure[3D view of the work shop]{%
\includegraphics[width=2.6in]{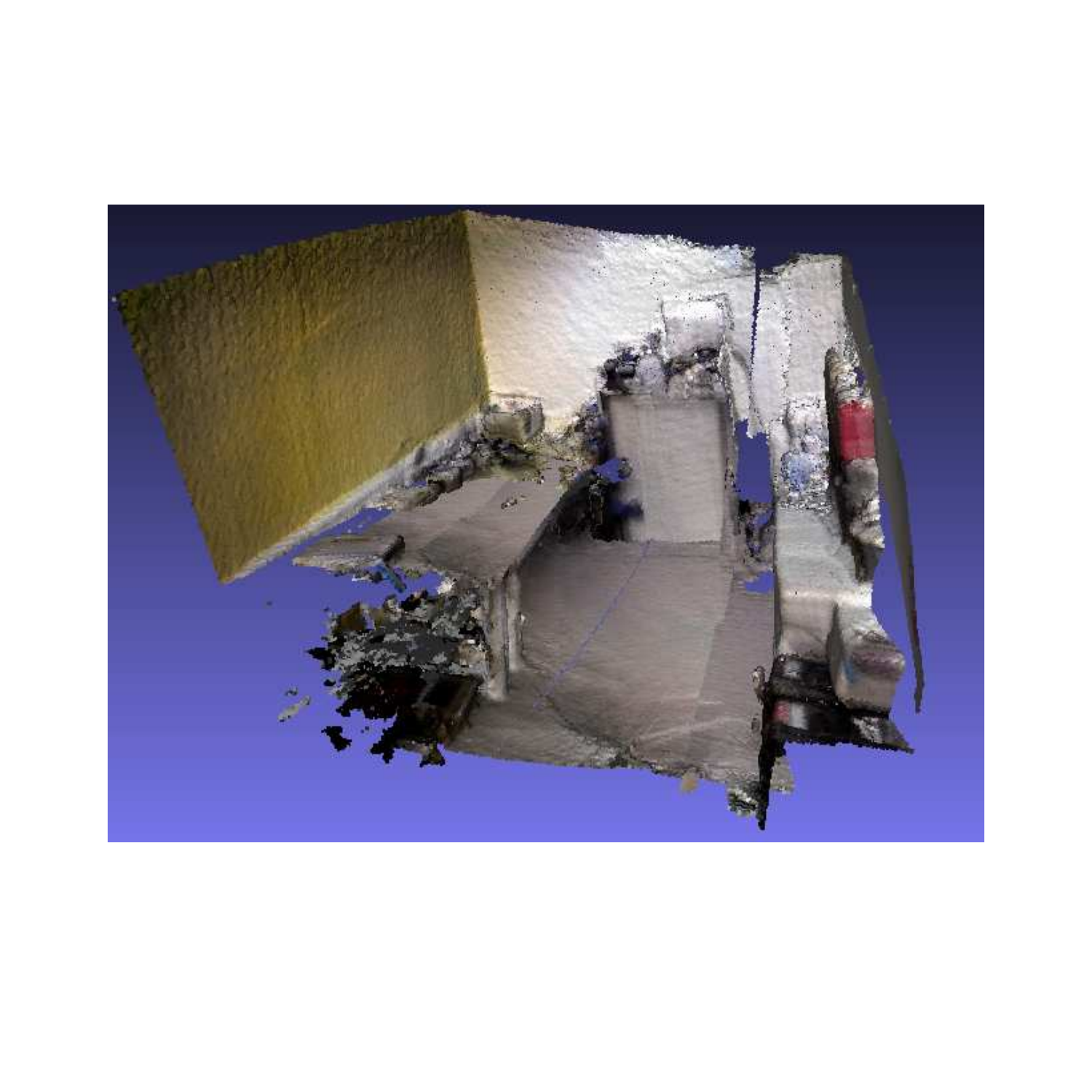}
\label{fig:room83d}}
\quad
\subfigure[Plan view with ground truth skeleton locations and their orientations]{%
\includegraphics[width=2.6in]{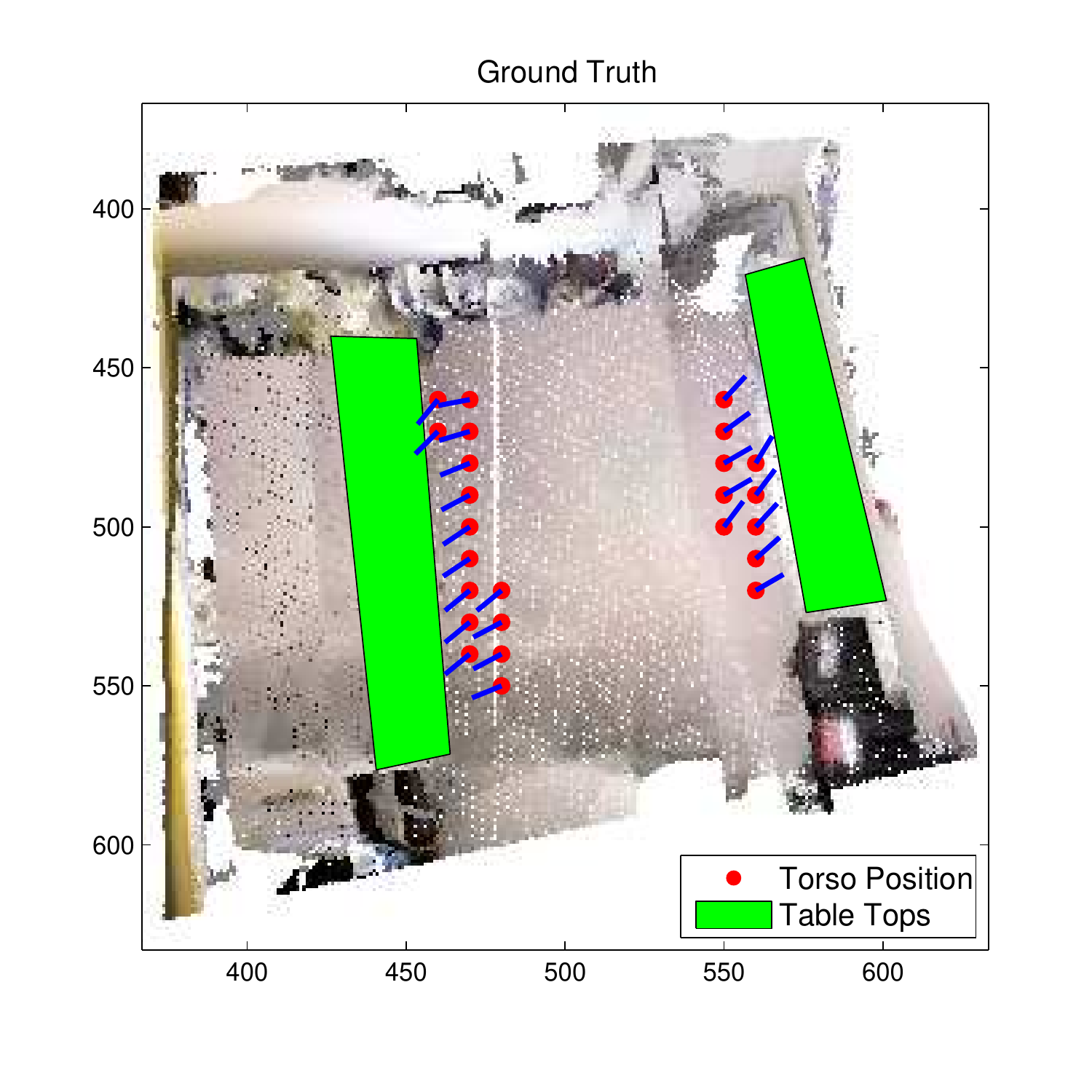}
\label{fig:room8_ground}}

\subfigure[S-SVM Classifier]{%
\includegraphics[width=0.3\textwidth]{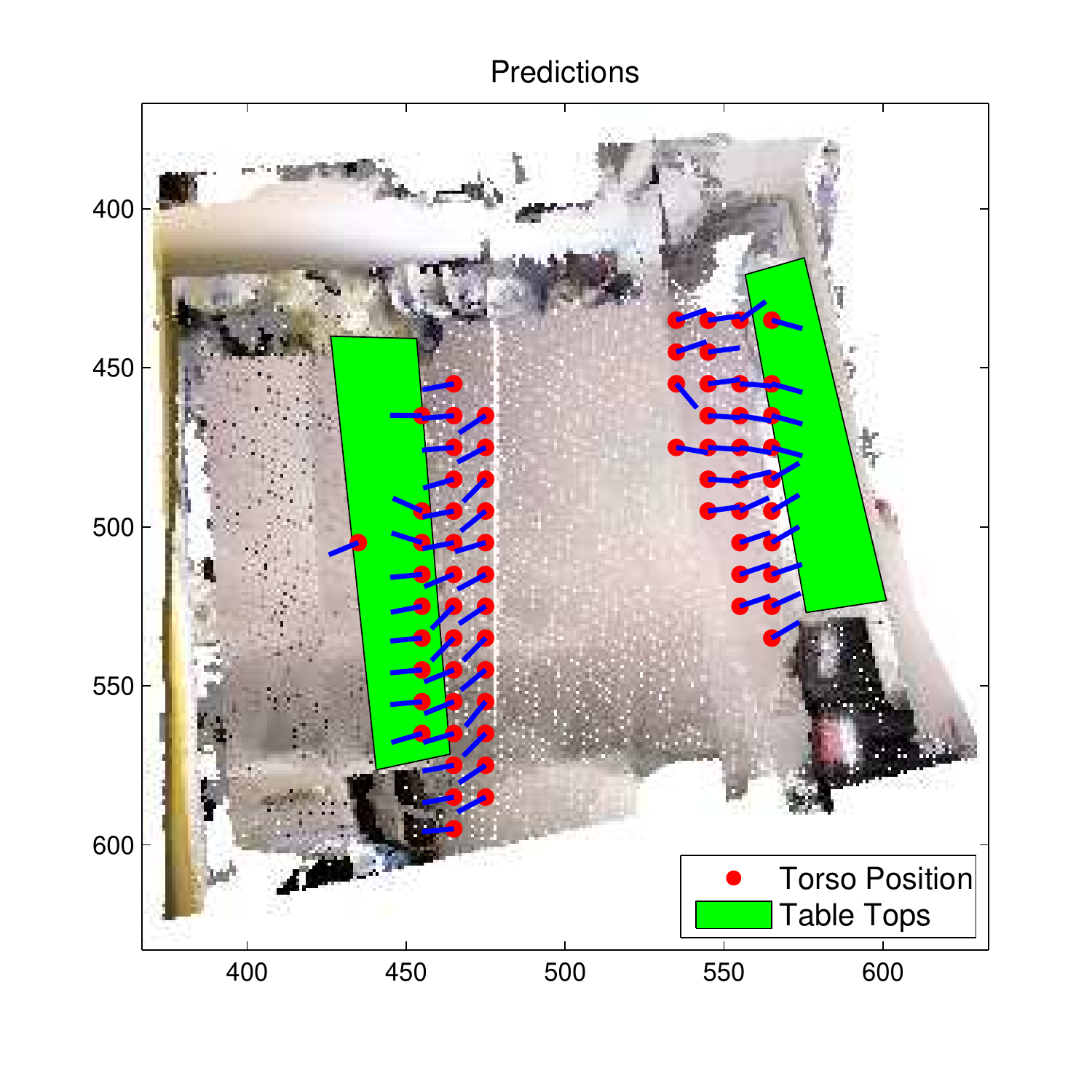}
\label{fig:prediction_leaning_SSVM}}
\subfigure[S-SVM Classifier]{\raisebox{0.0mm}{%
\includegraphics[width=0.3\textwidth]{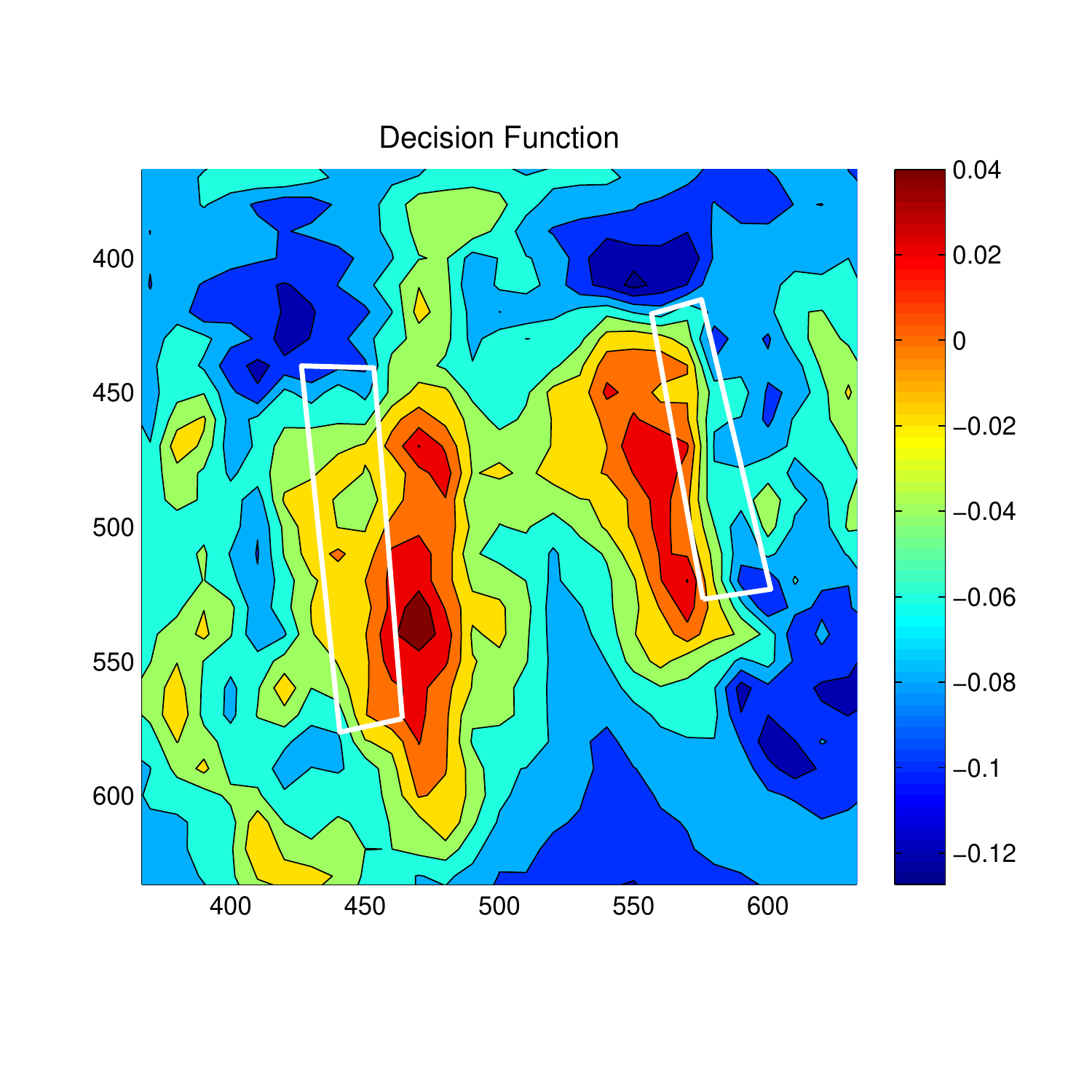}}
\label{fig:decision_leaning_SSVM}}
\subfigure[S-SVM Classifier]{\raisebox{10mm}{%
\includegraphics[width=0.3\textwidth]{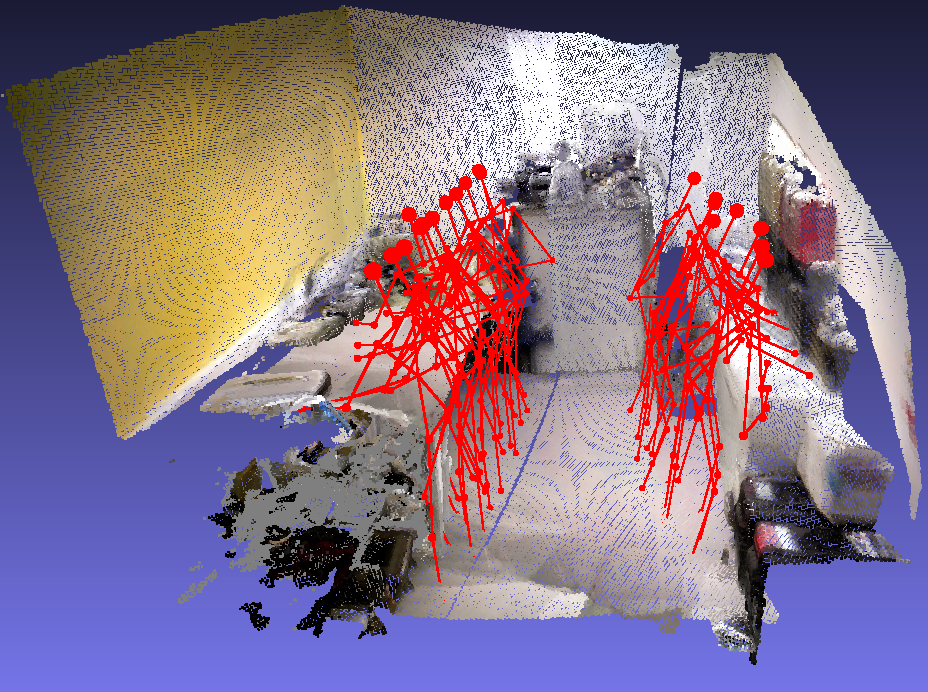}}
\label{fig:skeleton_leaning_SSVM}}

\subfigure[Focused-SVM Classifier]{%
\includegraphics[width=0.3\textwidth]{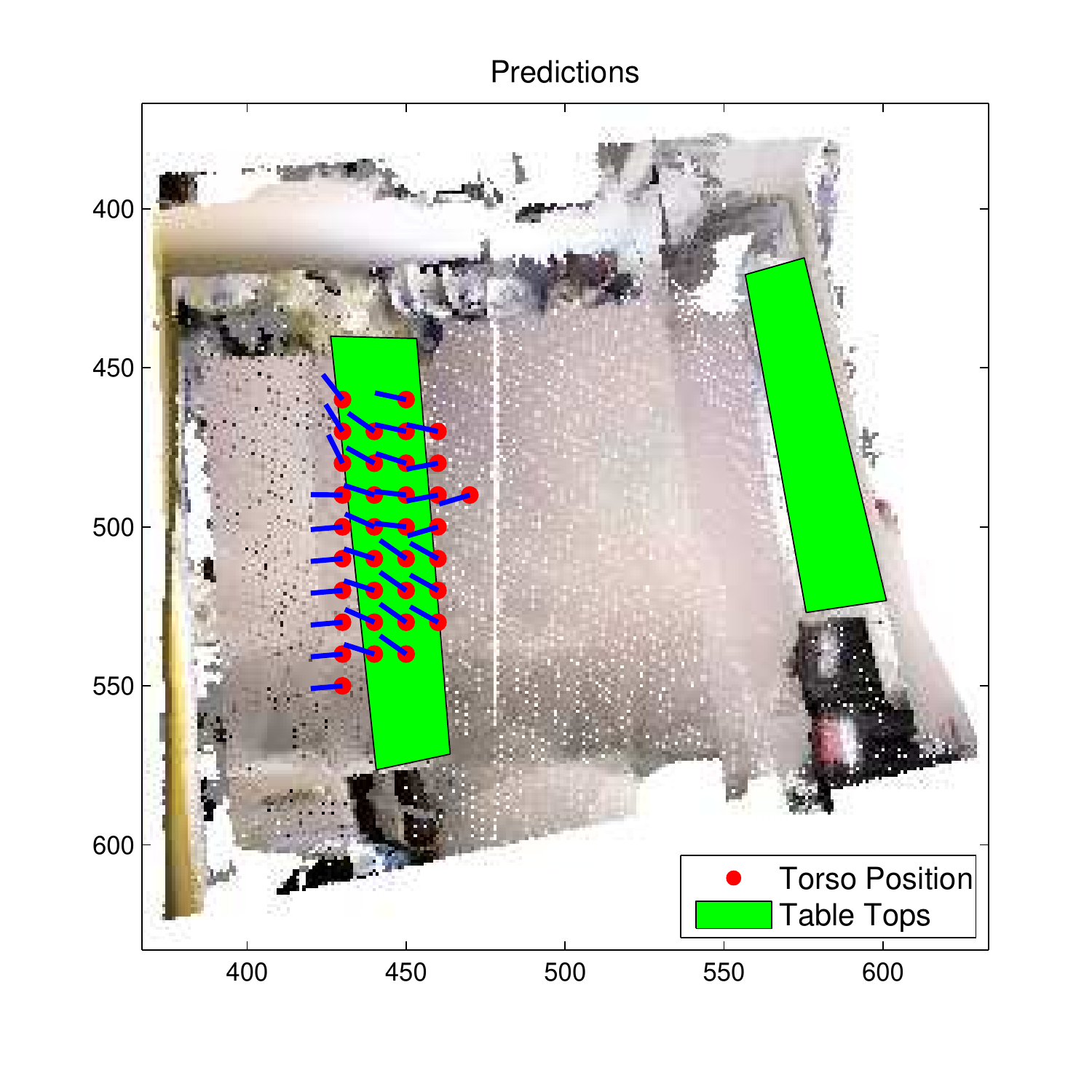}
\label{fig:prediction_leaning_focused}}
\subfigure[Focused-SVM Classifier]{%
\includegraphics[width=0.3\textwidth]{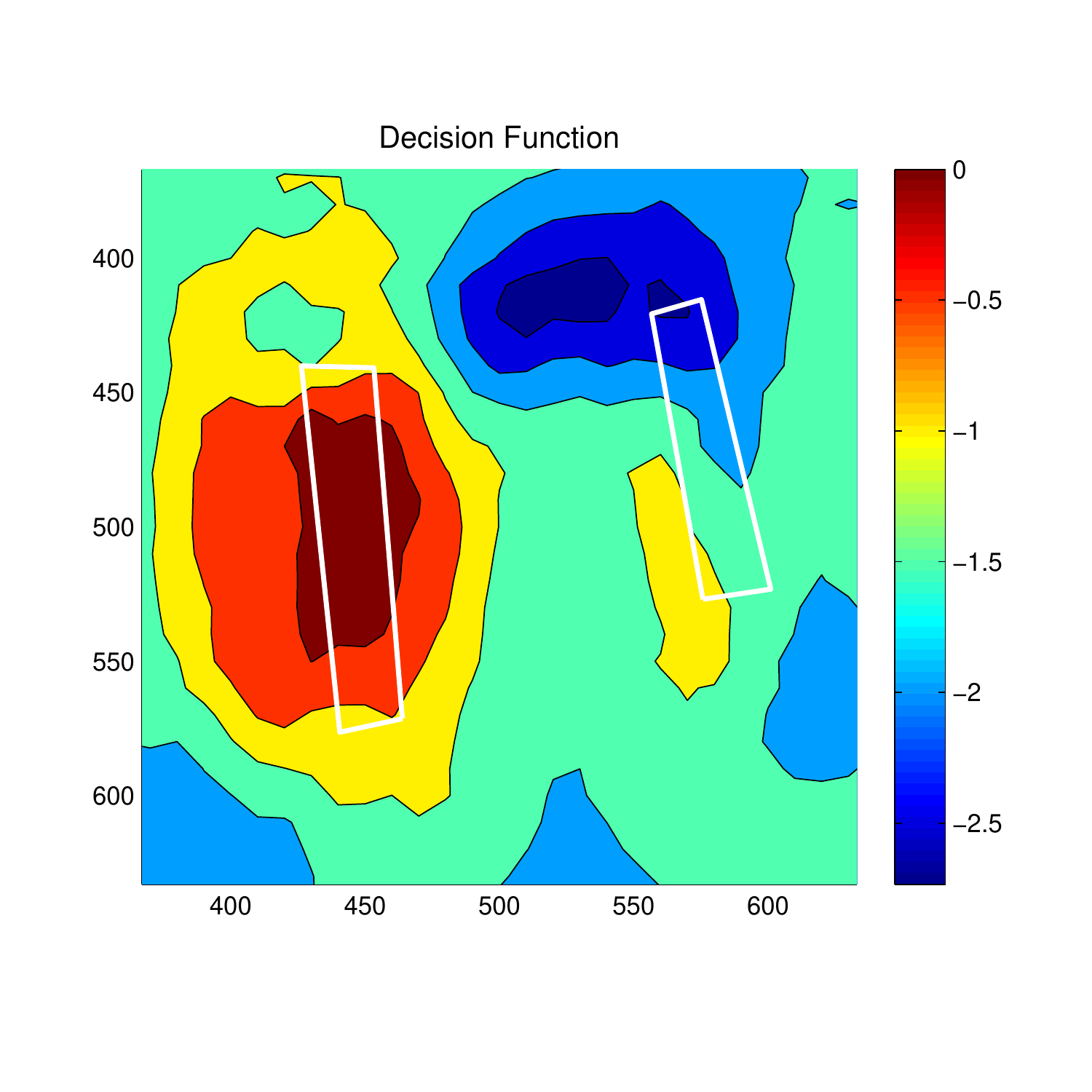}
\label{fig:decison_leaning_focused}}
\subfigure[Focused-SVM  Classifier]{%
\includegraphics[width=0.3\textwidth]{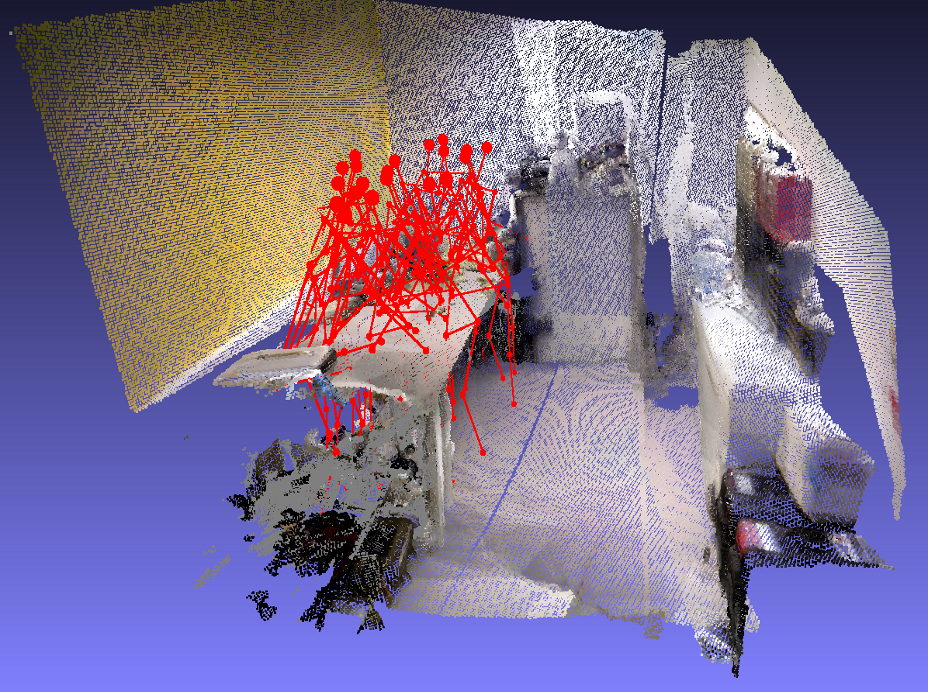}
\label{fig:skeleton_leaning_focused}}
\caption{Classification results of the Standing-Working Affordance}
\label{fig:prediction_standing_SSVM_all}
\end{figure*}

 Fig. \ref{fig:decision_working_SSVM} and \ref{fig:decision_working_DCM} shows the decision function values for the `sitting-working' affordance. Both S-SVM and DCM-SVM classifiers have predicted similar decision function values and have predicted high confidence values on locations where most of the chairs are located. Relatively,  high confidence values can be observed in both classifiers decision functions towards the bottom-left corner of the room. A careful examination reveals that those locations   consists of geometric features very similar to a chair (Right angled horizontal and vertical surfaces).

Fig. \ref{fig:skeleton_working_SSVM} and Fig. \ref{fig:skeleton_working_DCM}  shows the skeleton map for the affordance type Sitting-Working. In this case the DCM-SVM  classifier has marginally outperformed the S-SVM classifier. In particular S-SVM classifier has predicted a skeleton on the working desk towards the left side of the room and none of the classifiers  have predicted any skeleton on the chair at the top-left corner of the room. This may be because the chair is oriented away from the nearby desk, and therefore it could be argued, that chair is not supporting the Sitting-working affordance type. Overall, both classifiers have predicted a very similar skeleton map.

\subsubsection{Standing-Working Affordance}

Fig. \ref{fig:room83d} shows the room used to analyze the performances of the `standing-working affordance'. The  green color rectangle areas indicate workbenches of the lab space. According to the imbalance test results discussed in a previous section, the focused resampling SVM (Focused-SVM) method recorded the highest F1-score in the Standing-Working affordance. Therefore the results of the S-SVM method are compared with the Focus-SVM method.

The prediction results of the S-SVM classifier and Focused-SVM classifiers are shown in  
Fig. \ref{fig:prediction_leaning_SSVM} and\ref{fig:prediction_leaning_focused}  . It is clear from this figure that the S-SVM classifier has predicted better results than the Focused-SVM classifier. The Focused-SVM classifier has failed to predict any skeleton model at the workbench towards the right side of room. Although it has predicted some skeletons at the other workbench towards the right side of the room, most of them are located inside the workbench. In contrast, the S-SVM classifier has predicted skeletons at both workbenches and they are nicely oriented towards the working areas of the benches.

Fig. \ref{fig:decision_leaning_SSVM} and Fig. \ref{fig:decison_leaning_focused} show the values of the decision function for the affordance type Standing-Working. As indicated by the prediction results, the S-SVM classifier has placed high confidence values near the two workbenches whereas the Focused-SVM classifier has placed a number of  high confidence values inside the workbench towards the left of the room. Therefore, the  S-SVM classifier has predicted much better localized results than the Focused-SVM classifier.

Fig. \ref{fig:skeleton_leaning_SSVM} and Fig. \ref{fig:skeleton_leaning_focused}  show the skeleton map for the affordance type Standing-Working. In this example, the S-SVM method has predicted a very accurate skeleton map with skeletons located near the working benches with correct orientations. However, the Focused-SVM method has only predicted skeletons around one of the workbenches and some of the skeletons are located inside the workbench. In this case, S-SVM method has outperformed the Focus-SVM method.

\subsection{Quantitative Analysis}

Table \ref{Table_struct_summary} shows the comparison of average performance measures of k-fold cross validation between S-SVM and other methods. For the methods discussed previously, Table \ref{Table_struct_summary} shows the average  precision and recall values of the best F1-score recorded in class imbalance test for each affordance type. It is clear from this data the S-SVM method has outperformed other methods. 
The S-SVM method has recoded the highest F1-scores in the affordance types Sitting-Relaxing and Standing-Working by a considerable margin to the next best F1-score of  each category. It has recorded the third highest F1-score in Sitting-Working affordance  with only  -0.04 $\%$ difference to the highest recorded F1-score of that category. Overall, the S-SVM method has shown  consistent performance in all three affordance types. This is because, unlike other methods, S-SVM has been trained directly optimizing on F1-score. Therefore it is minimally affected by the class imbalance problem.

\begin{table}[]
\centering
\caption{Comparison of performance measures in the K-fold cross validation test. Structured SVM and Other Methods.}
\label{Table_struct_summary}
\begin{tabular}{|l|l|c|c|c|}
\hline
{\bf Affordance Type}              & {\bf SVM Algorithm}  & \multicolumn{1}{l|}{{\bf Precision}} & \multicolumn{1}{l|}{{\bf Recall}} & \multicolumn{1}{l|}{{\bf F1-Score}} \\ \hline
\multirow{6}{*}{Sitting- Relaxing} & Random Sampling      & 0.36                                 & 0.33                              & 0.35                                \\ \cline{2-5} 
                                   & Focused Sampling     & 0.32                                 & 0.37                              & 0.35                                \\ \cline{2-5} 
                                   & Z- SVM               & 0.51                                 & 0.35                              & 0.35                                \\ \cline{2-5} 
                                   & Different Cost Model & 0.28                                 & 0.44                              & 0.34                                \\ \cline{2-5} 
                                   & Structured SVM       & 0.65                                 & 0.80                              & {\bf 0.72}                          \\ \cline{2-5} 
                                   & \multicolumn{4}{l|}{}                                                                                                                 \\ \hline
\multirow{6}{*}{Standing-Working}  & Random Sampling      & 0.09                                 & 0.49                              & 0.13                                \\ \cline{2-5} 
                                   & Focused Sampling     & 0.17                                 & 0.30                              & 0.22                                \\ \cline{2-5} 
                                   & Z- SVM               & 0.12                                 & 0.31                              & 0.15                                \\ \cline{2-5} 
                                   & Different Cost Model & 0.10                                 & 0.42                              & 0.14                                \\ \cline{2-5} 
                                   & Structured SVM       & 0.28                                 & 0.81                              & {\bf 0.42}                          \\ \cline{2-5} 
                                   & \multicolumn{4}{l|}{}                                                                                                                 \\ \hline
\multirow{5}{*}{Sitting-Working}   & Random Sampling      & 0.55                                 & 0.52                              & 0.50                                \\ \cline{2-5} 
                                   & Focused Sampling     & 0.34                                 & 0.51                              & 0.38                                \\ \cline{2-5} 
                                   & Z- SVM               & 0.27                                 & 0.84                              & 0.37                                \\ \cline{2-5} 
                                   & Different Cost Model & 0.50                                 & 0.55                              & {\bf 0.52}                          \\ \cline{2-5} 
                                   & Structured SVM       & 0.40                                 & 0.77                              & 0.48                                \\ \hline
\end{tabular}
\end{table}

\section{Conclusions}

This paper addressed the problem of learning human context in indoor environments by
analyzing geometric features of the environment. To solve this problem, the concept
of affordance-map was proposed which predicts possible affordances in the environment
as human skeleton models with their affordance likelihoods. The problem of affordance
mapping was formulated as a multi label classification problem with a binary classifier
for each affordance type. In the first phase of this process, virtual human models were
moved across the grid locations of the 3D environment to capture features for classification.
These features were used to learn parameters of the classification model. Finally, learned
classifiers were used to infer the affordance-map in new unseen 3D environments.

To learn the spatial affordance map Support Vector Machine (SVM) was used and experimentally evaluated
evaluated. However, the SVM classifier recorded sub-optimum results due to the existence of class
imbalance in training data. In the search for a solution to this problem, three existing
SVM learners namely: focused re-sampling , zSVM and Different Cost model SVM were tested for learning affordances. These algorithms showed some tolerance to moderate class imbalances but failed to record acceptable results particularly in standing-working affordance. Therefore, this thesis proposed to use Structured SVM (S-SVM) optimized for F1-score to further improve affordance mapping results. This approach defined the affordance-map building process as a structured problem which could be learned efficiently even with extremely imbalanced data. The S-SVM method was experimentally analyzed 
both quantitatively and qualitatively and outperformed previously used classifiers for affordance mapping.

As furture work,the spatial  affordance map could be used to improve existing human activity recognition algorithms, social aware path planning and active object search algorithms \cite{piyathilakaaffordance,lasunmannedjournal,costsensitiveROBIO,piyathilaka2016affordance}. Most of the current activity recognition algorithms require a full body view of 
humans to recognize an activity correctly. However, this could become challenging in a
cluttered environment as body parts of humans could easily become occluded by objects
in the environments. In these scenarios, affordance-map could be used to predict locations
that give full views of humans so the robots can track human body parts more accurately.
On the other hand, information from an affordance-map could be used to improve human
detection and tracking algorithms. A similar approach used in this thesis for active object
search could be used to solve such problems.

\bibliographystyle{unsrt} 
\bibliography{root}

\begin{thebibliography}{10}

\bibitem{sarathSocial}
Stephan Sehestedt, Sarath Kodagoda, and Gamini Dissanayake.
\newblock Robot path planning in a social context.
\newblock In {\em Robotics Automation and Mechatronics (RAM), 2010 IEEE
  Conference on}, pages 206--211. IEEE, 2010.

\bibitem{FSR}
Lasitha Piyathilaka and Sarath Kodagoda.
\newblock Human activity recognition for domestic robots.
\newblock In {\em Field and Service Robotics Conference}, pages 567--572.
  Springer, 2013.

\bibitem{Gaussian}
Lasitha Piyathilaka and Sarath Kodaagoda.
\newblock Gaussian mixture based hmm for human daily activity recognition using
  3d skeleton features.
\newblock In {\em Industrial Electronics and Applications (ICIEA), 2013 8th
  IEEE Conference on}, pages 567--572. IEEE, 2013.

\bibitem{Hallucinated3D}
Yun Jiang, Hema Koppula, and Ashutosh Saxena.
\newblock Hallucinated humans as the hidden context for labeling 3d scenes.
\newblock In {\em Computer Vision and Pattern Recognition (CVPR), 2013 IEEE
  Conference on}, pages 2993--3000. IEEE, 2013.

\bibitem{Gibson}
Eleanor~Jack Gibson and Anne~D Pick.
\newblock {\em An ecological approach to perceptual learning and development}.
\newblock Oxford University Press, 2000.

\bibitem{ChairChair}
Helmut Grabner, Juergen Gall, and Luc~Van Gool.
\newblock What makes a chair a chair?
\newblock In {\em Computer Vision and Pattern Recognition (CVPR), 2011 IEEE
  Conference on}, pages 1529--1536. IEEE, 2011.

\bibitem{lasAffor}
Lasitha Piyathilaka and Sarath Kodagoda.
\newblock Active visual object search using affordance-map in real world : A
  human-centric approach.
\newblock In {\em ICARCV, The 13th International Conference on Control,
  Automation, Robotics and Vision}. IEEE, 2014.

\bibitem{Bang2008}
Sunlee Bang, Minho Kim, Sa-Kwang Song, and Soo-Jun Park.
\newblock {Toward real time detection of the basic living activity in home
  using a wearable sensor and smart home sensors}.
\newblock {\em Conference proceedings Annual International Conference of the
  IEEE Engineering in Medicine and Biology Society}, 2008:5200--5203, 2008.

\bibitem{Tapia2004}
Emmanuel~Munguia Tapia, Stephen~S Intille, and Kent Larson.
\newblock {Activity Recognition in the Home Using Simple and Ubiquitous
  Sensors}.
\newblock {\em Pervasive Computing}, 3001:158--175, 2004.

\bibitem{Lara_asurvey}
Duy T{\^a}m~Gilles Huynh.
\newblock {\em Human Activity Recognition with Wearable Sensors}.
\newblock PhD thesis, TU Darmstadt, September 2008.

\bibitem{Gupta2009}
Abhinav Gupta, Praveen Srinivasan, Jianbo Shi, and Larry~S Davis.
\newblock {Understanding videos, constructing plots learning a visually
  grounded storyline model from annotated videos}.
\newblock {\em 2009 IEEE Conference on Computer Vision and Pattern
  Recognition}, 21(4):2012--2019, 2009.

\bibitem{storkROMAN12}
Johannes Stork, Luciano Spinello, Jens Silva, Kai~O Arras, et~al.
\newblock Audio-based human activity recognition using non-markovian ensemble
  voting.
\newblock In {\em RO-MAN, 2012 IEEE}, pages 509--514. IEEE, 2012.

\bibitem{Kahn06whatis}
Peter~H. Kahn, Hiroshi Ishiguro, Batya Friedman, and Takayuki K.
\newblock What is a human? toward psychological benchmarks in the field of
  human robot interaction.
\newblock In {\em In Proceedings of the IEEE international}, pages 364--371,
  2006.

\bibitem{Bellotti93designfor}
Victoria Bellotti, H.~Distributed, Systems Architecture, and Csmil Technical.
\newblock Design for privacy in ubiquitous computing environments, 1993.

\bibitem{gibson2014ecological}
James~J Gibson.
\newblock {\em The Ecological Approach to Visual Perception: Classic Edition}.
\newblock Psychology Press, 2014.

\bibitem{jiang2012learning}
Yun Jiang, Marcus Lim, and Ashutosh Saxena.
\newblock Learning object arrangements in 3d scenes using human context.
\newblock {\em arXiv preprint arXiv:1206.6462}, 2012.

\bibitem{jiang2013hallucinated}
Yun Jiang, Hema Koppula, and Ankur Saxena.
\newblock Hallucinated humans as the hidden context for labeling 3d scenes.
\newblock In {\em Computer Vision and Pattern Recognition (CVPR), 2013 IEEE
  Conference on}, pages 2993--3000. IEEE, 2013.

\bibitem{c1}
Jaeyong Sung, Colin Ponce, Bart Selman, and Ashutosh Saxena.
\newblock Unstructured human activity detection from rgbd images.
\newblock In {\em Robotics and Automation (ICRA), 2012 IEEE International
  Conference on}, pages 842--849. IEEE, 2012.

\bibitem{ccny}
Ivan Dryanovski, Roberto~G Valenti, and Jizhong Xiao.
\newblock Fast visual odometry and mapping from rgb-d data.
\newblock In {\em Robotics and Automation (ICRA), 2013 IEEE International
  Conference on}, pages 2305--2310. IEEE, 2013.

\bibitem{cortes1995support}
Corinna Cortes and Vladimir Vapnik.
\newblock Support-vector networks.
\newblock {\em Machine learning}, 20(3):273--297, 1995.

\bibitem{joachims1999making}
Thorsten Joachims.
\newblock Making large scale svm learning practical.
\newblock Technical report, Universit{\"a}t Dortmund, 1999.

\bibitem{suykens1999least}
Johan~AK Suykens and Joos Vandewalle.
\newblock Least squares support vector machine classifiers.
\newblock {\em Neural processing letters}, 9(3):293--300, 1999.

\bibitem{tang2009svms}
Yuchun Tang, Yan-Qing Zhang, Nitesh~V Chawla, and Sven Krasser.
\newblock Svms modeling for highly imbalanced classification.
\newblock {\em Systems, Man, and Cybernetics, Part B: Cybernetics, IEEE
  Transactions on}, 39(1):281--288, 2009.

\bibitem{akbani2004applying}
Rehan Akbani, Stephen Kwek, and Nathalie Japkowicz.
\newblock Applying support vector machines to imbalanced datasets.
\newblock In {\em Machine Learning: ECML 2004}, pages 39--50. Springer, 2004.

\bibitem{wu2003adaptive}
Gang Wu and Edward~Y Chang.
\newblock Adaptive feature-space conformal transformation for imbalanced-data
  learning.
\newblock In {\em ICML}, pages 816--823, 2003.

\bibitem{chawla2002smote}
Nitesh~V. Chawla, Kevin~W. Bowyer, Lawrence~O. Hall, and W.~Philip Kegelmeyer.
\newblock Smote: synthetic minority over-sampling technique.
\newblock {\em Journal of artificial intelligence research}, pages 321--357,
  2002.

\bibitem{dehmeshki2004classification}
Jamshid Dehmeshki, Jun Chen, Manlio~Valdivieso Casique, and Mustafa Karakoy.
\newblock Classification of lung data by sampling and support vector machine.
\newblock In {\em Engineering in Medicine and Biology Society, 2004. IEMBS'04.
  26th Annual International Conference of the IEEE}, volume~2, pages
  3194--3197. IEEE, 2004.

\bibitem{he2009learning}
Haibo He, Edwardo Garcia, et~al.
\newblock Learning from imbalanced data.
\newblock {\em Knowledge and Data Engineering, IEEE Transactions on},
  21(9):1263--1284, 2009.

\bibitem{batuwita2010efficient}
Rukshan Batuwita and Vasile Palade.
\newblock Efficient resampling methods for training support vector machines
  with imbalanced datasets.
\newblock In {\em Neural Networks (IJCNN), The 2010 International Joint
  Conference on}, pages 1--8. IEEE, 2010.

\bibitem{imam2006z}
Tasadduq Imam, Kai~Ming Ting, and Joarder Kamruzzaman.
\newblock z-svm: an svm for improved classification of imbalanced data.
\newblock In {\em AI 2006: Advances in Artificial Intelligence}, pages
  264--273. Springer, 2006.

\bibitem{joachims1999transductive}
Thorsten Joachims.
\newblock Transductive inference for text classification using support vector
  machines.
\newblock In {\em ICML}, volume~99, pages 200--209, 1999.

\bibitem{li2009landmark}
Yunpeng Li, David~J Crandall, and Daniel~P Huttenlocher.
\newblock Landmark classification in large-scale image collections.
\newblock In {\em Computer vision, 2009 IEEE 12th international conference on},
  pages 1957--1964. IEEE, 2009.

\bibitem{joachims2005support}
Thorsten Joachims.
\newblock A support vector method for multivariate performance measures.
\newblock In {\em Proceedings of the 22nd international conference on Machine
  learning}, pages 377--384. ACM, 2005.

\bibitem{tsochantaridis2004support}
Ioannis Tsochantaridis, Thomas Hofmann, Thorsten Joachims, and Yasemin Altun.
\newblock Support vector machine learning for interdependent and structured
  output spaces.
\newblock In {\em Proceedings of the twenty-first international conference on
  Machine learning}, page 104. ACM, 2004.

\bibitem{piyathilakaaffordance}
Lasitha Piyathilaka and Sarath Kodagoda.
\newblock Affordance-map: A map for context-aware path planning.
\newblock In {\em ACRA, Australasian Conference on Robotics and Automation
  2014}. ARAA, 2014.

\bibitem{lasunmannedjournal}
Lasitha Piyathilaka and Sarath Kodagoda.
\newblock Learning hidden human conetext in 3d office scenes by mapping
  affordances through virtual humans.
\newblock {\em Unmanned Systems}, 2015.

\bibitem{costsensitiveROBIO}
L.~{Piyathilaka} and S.~{Kodagoda}.
\newblock Affordance-map: Mapping human context in 3d scenes using
  cost-sensitive svm and virtual human models.
\newblock In {\em 2015 IEEE International Conference on Robotics and
  Biomimetics (ROBIO)}, pages 2035--2040, Dec 2015.

\bibitem{piyathilaka2016affordance}
Jayaweera Mudiyanselage Lasitha~Chandana Piyathilaka.
\newblock {\em Affordance-map: learning hidden human context in 3D scenes
  through virtual human models}.
\newblock PhD thesis, 2016.

\end{thebibliography}

\end{document}